\documentclass[9.5pt,journal,finalsubmission,twocolumn]{IEEEtran}
\pdfoutput=1

\usepackage{graphicx}
\usepackage{amsmath,amsfonts,amssymb,amscd,amsthm,xspace}
\usepackage{import}
\usepackage{subfig}
\usepackage{xcolor}
\usepackage{mathtools}
\usepackage{math}
\usepackage{enumitem}
\usepackage{multirow}
\usepackage{booktabs}
\usepackage{url}
\usepackage[lined,boxed,commentsnumbered,ruled,norelsize]{algorithm2e}

\usepackage[final]{review}
\setrevision{3}

\newcommand{\eref}[1]{~(\ref{#1})}
\newcommand{\sref}[1]{Section~\ref{#1}}

\newcommand{\fref}[1]{Fig.~\ref{#1}}
\newcommand{\tref}[1]{Table~\ref{#1}}
\newcommand{\besta}[1]{$\underline{\mathbf{#1}}$}
\newcommand{\bestb}[1]{$\mathbf{#1}$}
\newcommand{\ir}[1][]{^{#1(r)}}
\newcommand{\irp}[1][]{^{#1(r+1)}}
\newcommand{\E}{\textrm{E}}
\newcommand{\tr}{^\top}
\newcommand{\mah}[2]{\left\|#1\right\|_{#2}}
\renewcommand{\det}[1]{\left|#1\right|}

\newcommand{\argmin}{\operatornamewithlimits{argmin}}
\newcommand{\up}{\operatornamewithlimits{high}}
\newcommand{\low}{\operatornamewithlimits{low}}
\newcommand{\eg}{e.g. }
\newcommand{\ie}{i.e. }

\DeclarePairedDelimiter\norm{\lVert}{\rVert}%
\DeclarePairedDelimiter\abs{\lvert}{\rvert}%

\makeatletter
\newcommand{\removelatexerror}{\let\@latex@error\@gobble}
\makeatother

\title{EM Algorithms for Weighted-Data Clustering\\
with Application to Audio-Visual Scene Analysis}
\author{Israel D. Gebru, Xavier Alameda-Pineda, Florence Forbes and Radu Horaud
\IEEEcompsocitemizethanks{\IEEEcompsocthanksitem I.~D. Gebru, F. Forbes and R. Horaud, INRIA Grenoble Rh\^one-Alpes, Montbonnot Saint-Martin, France. \protect\\
E-mail: \{israel-$ $-dejene.gebru, florence.forbes, radu.horaud\}@inria.fr
\IEEEcompsocthanksitem X. Alameda-Pineda, INRIA Grenoble Rh\^one-Alpes and University of Trento, Trento, Italy. Email: 
xavier.alamedapineda@unitn.it.}
\thanks{Funding from the European Union FP7 ERC Advanced Grant  VHIA (\#340113) and STREP EARS (\#609645) is greatly acknowledged.}
\thanks{F. Forbes and R. Horaud are partially supported by a XEROX University Affairs Committee (UAC) grant (2015-2017).}
}

\begin{document}
\maketitle

\begin{abstract}
Data clustering has received a lot of attention and numerous methods, algorithms and software packages are available. Among these techniques, 
parametric finite-mixture models play a central role due to their interesting mathematical properties and to the existence of maximum-likelihood 
estimators based on expectation-maximization (EM). In this paper we propose a new mixture model that associates a weight with each observed point. We 
introduce the weighted-data Gaussian mixture and we derive two EM algorithms. The first one considers a fixed weight for each observation. The second 
one treats each weight as a random variable following a gamma distribution. We propose a model selection method based on a minimum message length 
criterion, provide a weight initialization strategy, and validate the proposed algorithms by comparing them with several state of the art parametric 
and non-parametric clustering techniques. We also demonstrate the effectiveness and robustness of the proposed clustering technique in the presence 
of heterogeneous data, namely audio-visual scene analysis.
\end{abstract}
\section{Introduction}
\label{sec:intro}

Finding significant groups in a set of data points is a central problem in many fields. Consequently, clustering has received a lot of attention, and many methods, algorithms and software packages are available today. Among these techniques, parametric finite mixture models play a paramount role, due to their interesting mathematical properties as well as to the existence of maximum likelihood estimators based on expectation-maximization (EM) algorithms. While the finite Gaussian mixture (GMM) \cite{McLachlanPeel2000} is the model of choice, it is extremely sensitive to the presence of outliers. Alternative robust models have been proposed in the statistical literature, such as mixtures of t-distributions \cite{McLachlanPeel2000b} and their numerous variants, \eg \cite{Bishop2005,Archambeau2007,sun2010robust,Andrews2012,ForbesWraith2014,lee2014finite}. In contrast to the Gaussian case, no closed-form solution exists for the t-distribution and tractability is maintained via the use of EM and a Gaussian scale mixture representation,
$
\mathcal{T}(\xvect | \muvect, \Sigmamat, \alpha) = \int_{0}^{\infty} \mathcal{N}(\xvect | \muvect, \Sigmamat/w) \;  \mathcal{G}(w, \alpha/2, \alpha/2) dw,
$
where $\xvect$ is an observed vector, $\mathcal{N}$ is the multivariate Gaussian distribution with mean $\muvect$ and covariance $\Sigmamat/w$, and $\mathcal{G}$ is the gamma distribution of a univariate positive variable $w$ parameterized by $\alpha$. \addnote[mixture]{1}{In the case of mixtures of t-distributions, with mixing coefficients $\pi_k$, 
$\sum_{k=1}^K\pi_k \mathcal{T}(\xvect | \muvect_k, \Sigmamat_k, \alpha_k)$, a latent variable $w$ can also be introduced. Its distribution is a mixture of $K$ gamma distributions that accounts for the component-dependent $\alpha_k$ \cite{McLachlanPeel2000b}. Clustering is then usually performed associating a positive variable $w_i$, distributed as $w$, with \textbf{each} observed point $\xvect_i$. The distributions of both $w_i$ and $\xvect_i$ do not depend on $i$.  The observed data are drawn from i.i.d. variables, distributed according to the t-mixture, or one of its variants \cite{McLachlanPeel2000b,Bishop2005,Archambeau2007,sun2010robust,Andrews2012,ForbesWraith2014,lee2014finite}.}

\addnote[not-iid]{1}{
In this paper we propose a finite mixture model in which  variable $w_i$  is used as a weight to account for the reliability of the observed $\xvect_i$ and this independently on its assigned cluster. The distribution of $w_i$ is not a gamma mixture anymore but has to depend on $i$ to allow each data point to be potentially treated differently.  
 In contrast to mixtures of t-distributions, it follows that the observed data are independent \textit{but not} identically distributed. 
}
We introduce the weighted-data Gaussian mixture model (WD-GMM). We distinguish two cases, namely (i)~the weights are known a priori and hence they 
are fixed, and (ii)~the weights are modeled as variables and hence they are iteratively updated, given initial estimates. We show that in the case of 
fixed weights, the GMM parameters can be estimated via an extension of the standard EM which will be referred to as the \textit{fixed 
weighted-data} EM algorithm (FWD-EM). 
Then we consider the more general case of weights that are treated as random variables. We model these variables with gamma distributions (one distribution for each variable) and we formally derive a closed-form EM algorithm which will be referred to as the \textit{weighted-data} EM algorithm (WD-EM). While the M-step of the latter is similar to the M-step of FWD-EM, the E-step is considerably different as both the posterior probabilities (responsibilities) and the parameters of the posterior gamma distributions (the weights) are updated (E-Z-step and E-W-step).
\addnote[pearson]{1}{
The responsibilities are computed using the Pearson type~VII distribution (the reader is referred to \cite{sun2010robust} for a recent 
discussion regarding this distribution), also called the  Arellano-Valle and Bolfarine generalized t-distribution \cite{KotzNadarajah2004}, and 
the parameters of the posterior gamma distributions are computed from the prior gamma parameters and from the Mahalanobis distance between the data and the 
mixture means. 
Note that the weights play a different role than the responsibilities. Unlike the responsibilities, which are probabilities, the weights are random variables that can take arbitrary positive values. Their posterior means can be used as an absolute measure of the relevance of the data. Typically, an outlying data point which is far from any cluster center will have a small weight  while it may still be assigned with a significant responsibility value to the closest cluster. Responsibilities indicate which cluster center is the closest but not 
if any of them is close at all. 
}

The idea of weighted-data clustering has already been proposed in the framework of non-parametric clustering methods such as $K$-means and spectral 
clustering, \eg \cite{long2006spectral,tseng2007penalized,ackerman2012weighted,feldman2012data}. These methods generally propose to incorporate 
prior information in the clustering process in order to prohibit atypical data (outliers) to contaminate the clusters.
The idea of modeling data weights as random variables and to estimate them via EM was proposed in \cite{forbes2010weighted} in the particular framework of  Markovian brain image segmentation. In \cite{forbes2010weighted} it is shown that specific expert knowledge is not needed and that the data-weight distribution guide the model towards a satisfactory segmentation. A variational EM is proposed in \cite{forbes2010weighted} as their formulation has no closed form. In this paper we build on the idea that, instead of relying on prior information about atypical data, \eg \cite{long2006spectral,tseng2007penalized,ackerman2012weighted,feldman2012data}, we devise a novel EM algorithm that updates the weight distributions. 
\addnote[robust-explanation]{1}{The proposed method belongs to the \textit{robust clustering} category of mixture models because observed data that 
are far away from the cluster centers have little influence on the estimation of the means and covariances. 
}

\addnote[model-selection-intro]{1}{An important feature of mixture based clustering methods is to perform model selection on the premise that the 
number of components $K$ in the mixture corresponds to the number of clusters in the data. Traditionally, model selection is performed by obtaining a 
set of candidate models for a range of values of $K$ (assuming that the true value is in this range). The number of components is selected by 
minimizing a model selection criteria, such as the Bayesian inference criterion (BIC), minimum message length (MML), Akaike's information criteria 
(AIC) to cite just a few \cite{McLachlanPeel2000,figueiredo2002unsupervised}. The disadvantage of these methods is twofold. Firstly, a whole set of 
candidates has to be obtained and problems associated with running EM many times may emerge. Secondly, they provide a number of components that 
optimally approximate the density and not the true number of clusters present in the data. More recently, there seems to be a consensus among mixture 
model practitioners that a well-founded and computationally efficient model selection strategy is to start with a large number of components and to 
merge them \cite{hennig2010methods}. \cite{figueiredo2002unsupervised} proposes a practical algorithm that starts with a very large number of 
components (thus making the algorithm robust to initialization), iteratively annihilates components, redistributes the observations to the other 
components, and terminates based on the MML criterion. 
\cite{baudry2010combining} starts with an overestimated number of components using BIC, and then merges them hierarchically according to an entropy criterion. More recently \cite{melnykov2014merging} proposes a similar method that merges components based on measuring their pair-wise overlap.
}

Another trend in handling the issue of finding the proper number of components is to consider Bayesian non-parametric mixture models. This allows the 
implementation of mixture models with an infinite number of components via the use of Dirichlet process mixture models. In 
\cite{rasmussen1999infinite,Gorur2010} an infinite Gaussian mixture (IGMM) is presented with a computationally intensive Markov Chain Monte Carlo 
implementation. \addnote[igmm]{3}{
At first glance, IGMM may appear similar to FWD-EM. However, these two algorithms are quite different. 
While IGMM is fully Bayesian the proposed FWD-EM is not, in the sense that no priors are assumed on the parameters, typically the means and 
covariance matrices. IGMM implies Student predictive distributions while FWD-EM involves only Gaussian distributions.}

More recently, more flexibility in the cluster shapes has been allowed by considering infinite mixture of infinite Gaussian mixtures 
(I$^2$GMM) \cite{yerebakan2014infinite}. The flexibility is however limited to a cluster composed of sub-clusters of identical shapes and 
orientations, which may alter the performance of this approach. Altogether, IGMM and I$^2$GMM are not designed to handle outliers, as illustrated in 
Section \ref{sec:experiments}, Figs. \ref{fig:sim_mixed}-f and \ref{fig:sim_mixed}-g. Infinite Student mixture models have also been considered 
\cite{Wei2012}, but inference requires a variational Bayes approximation which generates additional computational complexity.

Bayesian non-parametrics, although promising techniques, require a fully Bayesian setting. The latter, however, induces additional complexity for 
handling priors and hyper-priors,  especially in a multi-variate context. In contrast, our latent variable approach allows exact inference. With 
respect to model selection, we therefore propose to extend  the method  of \cite{figueiredo2002unsupervised} to weighted-data mixtures. We formally 
derive an MML criterion for the weighted-data mixture model and we plug this criterion into an efficient algorithm which, starting with a large number 
of components, simultaneously estimates the model parameters, the posterior probabilities of the weights and the optimal number of components.

\addnote[av-intro]{1}{
We also propose to apply the proposed weighted-data robust clustering method to the problem of fusing auditory and visual information. 
This problem arises when the task is, \eg to detect a person that is both seen and heard, such as an active speaker. Single-modality signals -- 
vision-only or audio-only -- are often either weak or ambiguous, and it may be useful to combine information from different 
sensors, \eg cameras and microphones. There are several difficulties associated with audio-visual fusion from a data clustering perspective: the 
two sensorial modalities (i)~live in different spaces, (ii)~are contaminated by different types of noise with different distributions, (iii)~have 
different spatiotemporal distributions, and (iv)~are perturbed by different physical phenomena, \eg acoustic reverberations, lighting conditions, 
etc. For example, a speaker may face the camera while he/she is silent and may emit speech while he/she turns his/her face away from the camera. 
Speech signals have 
sparse spectro-temporal structure and they are mixed with other sound sources, such as music or background noise. Speaker faces may be totally or partially occluded, in which case face detection and localization is extremely unreliable. We show that the proposed method is well suited to find audio-visual clusters and to discriminate between speaking and silent people.
}


The remainder of this paper is organized as follows. \sref{sec:mm_wd} outlines the weighted-data mixture model; \sref{sec:EM-fixed} sketches the 
FWD-EM algorithm. Weights modeled with random variables are introduced in \sref{sec:weight_description} and the  
WD-EM is described in detail in \sref{sec:em_formulation}. \sref{sec:model-selection} details how to deal with an unknown number of clusters and 
\sref{sec:initialization} addresses the issue of algorithm initialization. In \sref{sec:experiments} the proposed algorithms are tested and compared with several other parametric and non-parametric
clustering methods. \sref{sec:avexperiment} addresses clustering of audio-visual data. \sref{sec:conclusion} concludes the paper. Additional results and videos are available online.\footnote{\url{https://team.inria.fr/perception/research/wdgmm/}}


\section{Gaussian Mixture with Weighted Data}
\label{sec:mm_wd}
In this Section, we present the intuition and the formal definition of the proposed weighted-data model. Let $\xvect\in\mathbb{R}^d$ be a random vector following a multivariate Gaussian distribution with mean
$\muvect\in\mathbb{R}^d$ and covariance $\Sigmamat\in\mathbb{R}^{d\times d}$, namely $p(\xvect|\thetavect)={\mathcal N} \left(\xvect;
\muvect,\Sigmamat\right)$, with the notation $\thetavect=\{\muvect,\Sigmamat\}$. Let $w>0$ be a weight indicating the relevance of the observation $\xvect$. 
Intuitively, higher the weight $w$, stronger the impact of $\xvect$. The weight can therefore be incorporated into the model
by ``\textit{observing $\xvect$ $w$ times}''. In terms of the likelihood function, this is equivalent to raise
$p(\xvect ; \thetavect)$ to the power $w$, \ie $\mathcal{N}(\xvect;\muvect,\Sigmamat)^w$. However, the latter is
not a probability distribution since it does not integrate to one. It is straightforward to notice that
$
\mathcal{N}(\xvect;\muvect,\Sigmamat)^w\propto \mathcal{N}(\xvect;\muvect,\Sigmamat / w)
$.
\addnote[precision]{1}{Therefore, $w$ plays the role of the precision
and  is different 
for each datum $\xvect$. }
Subsequently, we write:
\begin{equation}
\hat{p}(\xvect ; \thetavect,w) =  \mathcal{N}\left(\xvect;\muvect,\frac{1}{w}\Sigmamat\right),
\label{eq:normalscaled}
\end{equation}
from which we derive a mixture model with $K$ components:
\begin{equation}
 \tilde{p}(\xvect ; \Thetavect,w) 
= \sum_{k=1}^K \pi_k\, \mathcal{N}\left(\xvect;\muvect_k,\frac{1}{w}\Sigmamat_k\right),
\label{eq:wdgmm}
\end{equation}
where $\Thetavect=\left\{\pi_1,\ldots,\pi_K,\thetavect_1,\ldots,\thetavect_K\right\}$ are the mixture parameters, $\pi_1,\ldots,\pi_K$ are the 
mixture coefficients satisfying $\pi_k\geq0$ and $\sum_{k=1}^K \pi_k=1$, $\thetavect_k=\{\muvect_k,\Sigmamat_k\}$ are the parameters of the 
$k$-th component and $K$ is the number of components. We will refer to the model in\eref{eq:wdgmm} as the \textit{weighted-data Gaussian mixture 
model} (WD-GMM). Let $\Xvect=\{\xvect_1,\ldots,\xvect_n\}$ be the observed data and $\Wvect=\{w_1, \ldots, w_n\}$ be the weights associated with 
$\Xvect$. We assume each $\xvect_i$ is independently drawn 
from\eref{eq:wdgmm} with $w=w_i$.
 The observed-data log-likelihood is:
\begin{equation}
 \ln \tilde{p}(\Xvect ; \Thetavect,\Wvect) 
\!=\! \sum_{i=1}^n \ln \! \left( \sum_{k=1}^K \pi_k\,\mathcal{N}\!\left(\xvect_i;\muvect_k,\frac{1}{w_i}\Sigmamat_k\right)\!\!\right). 
\label{eq:ll-wdgmm}%
\end{equation}
It is well known that direct maximization of the log-likelihood function is problematic in case of mixtures and that the expected complete-data log-likelihood must be considered instead. Hence, we 
introduce a set of $n$ hidden (assignment) variables
$\Zvect=\{z_1,\ldots,z_n\}$ associated with the observed variables $\Xvect$ and such that $z_i=k$, $k\in\{1,\ldots,K\}$ if and only if 
$\xvect_i$ is
generated by the $k$-th component of the mixture. In the following we first consider a fixed (given) number of mixture components $K$, 
we then extend the model to an unknown $K$, thus estimating the number of components from the data.

\section{EM with Fixed Weights}
\label{sec:EM-fixed}
The simplest case is when the weight values are provided at algorithm initialization, either using some prior knowledge or estimated from the 
observations (\eg Section~\ref{sec:initialization}), and are then kept fixed while alternating between the expectation and maximization steps. In 
this case, the expected complete-data log-likelihood is:
\begin{equation}
\mathcal{Q}_c\big(\Thetavect,\Thetavect\ir\big) = \E_{ P(\Zvect|\Xvect ; \Wvect, \Thetavect\ir)}\left[\ln P\left( \Xvect,\Zvect ; 
\Wvect,\Thetavect \right)\right],
\label{eq:q_c_function}
\end{equation}
where $\E_P[\cdot]$ denotes the expectation with respect to the distribution $P$. The $(r+1)$-th EM iteration consists of two
steps namely, the evaluation of the posterior distribution given the current model parameters $\Thetavect^{(r)}$ and the weights $\Wvect$ (E-step), and the maximization of\eref{eq:q_c_function} 
with respect to $\Thetavect$ (M-step):
\begin{equation}
\Thetavect\irp = \arg \max_{\Thetavect} \mathcal{Q}_c\big(\Thetavect,\Thetavect\ir\big).
\end{equation}
It is straightforward to show that this yields the following FWD-EM algorithm:
\subsection{The E-Step} The posteriors $ \eta_{ik}\irp =p(z_i = k  |  \xvect_i; w_i, \Thetavect\ir)$ are updated with:
\begin{equation}
 \eta_{ik}\irp =
\frac{\pi_k\ir \hat{p}(\xvect_i ; \thetavect_k\ir,w_i)}{\tilde{p}(\xvect_i ; \Thetavect\ir,
w_i)},
\label{eq:eta_update}
\end{equation}
\label{eq:fixed-posteriors}
where $\hat{p}$ and $\tilde{p}$ are defined in\eref{eq:normalscaled} and\eref{eq:wdgmm}.

\subsection{The M-Step} Expanding\eref{eq:q_c_function} we get:
\begin{align}
\label{eq:fixed-mstep}
 \mathcal{Q}_c\big(\Thetavect,\Thetavect\ir\big) =& \sum_{i=1}^n\sum_{k=1}^K \eta_{ik}\irp\ln \pi_k
\mathcal{N}\big(\xvect_i;\muvect_k;\frac{1}{w_i}\Sigmamat_k\big) \nonumber \\
 \stackrel{\Thetavect}{=}&\sum_{i=1}^n\sum_{k=1}^K\eta_{ik}\irp\left(\ln\pi_k - \ln\det{\Sigmamat_k}^{1/2}\right. \nonumber \\
- &\left. \frac{w_i}{2}(\xvect_i-\muvect_k)\tr\Sigmamat_k^{-1}(\xvect_i-\muvect_k)\right),
\end{align}
where $\stackrel{\Thetavect}{=}$ denotes equality up to a constant that does not depend on $\Thetavect$. By canceling out the derivatives with respect to the model parameters, we obtain the 
following update formulae for the mixture proportions, means, and covariances matrices:
\begin{equation}
\label{eq:prior-fixed}
 \pi_k\irp =\frac{1}{n}\sum_{i=1}^n\eta_{ik}\irp,
 \end{equation}
 \begin{equation}
\label{eq:mean-fixed}
 \muvect_k\irp = \frac{\displaystyle\sum_{i=1}^n w_i\eta_{ik}\irp\xvect_i}{\displaystyle\sum_{i=1}^n w_i\eta_{ik}\irp},
 \end{equation}
 \begin{equation}
\label{eq:cov-fixed}
\Sigmamat_k\irp = \frac{\displaystyle\sum_{i=1}^n w_i \eta_{ik}\irp 
\left(\xvect_i-\muvect_k\irp\right)\left(\xvect_i-\muvect_k\irp\right)\tr}{\displaystyle\sum_{i=1}^n \eta_{ik}\irp}.
\end{equation}

\section{Modeling the Weights}
\label{sec:weight_description}
As we already remarked, the weights play the role of precisions. The notable difference between standard finite mixture models and the proposed model is that there is a different weight $w_i$, hence a 
 different precision, associated with \textit{each} observation $\xvect_i$. Within a Bayesian formalism, the weights $\Wvect$ may be treated as random variables, rather than being fixed in advance, as in the previous case. Since\eref{eq:normalscaled} is a Gaussian, a convenient choice for the prior on $w$, $p(w)$ is the conjugate prior of the precision with known mean, \ie a gamma 
distribution. This ensures that the weight posteriors are gamma distributions as well. Summarizing we have:
\begin{align}
P(w;\phivect) =  \mathcal{G}\left(w;\alpha,\beta\right)
\label{eq:w_gamma}
= \Gamma\left(\alpha\right)\inverse \beta^\alpha w^{\alpha-1} e^{-\beta w},
\end{align}
where $\mathcal{G}\left(w;\alpha,\beta\right)$
is the gamma distribution, $\Gamma(\alpha) =\int_0^\infty t^{\alpha-1}e^{-t}
\textrm{d}t$ is the gamma function, and $\phivect=\{\alpha,\beta\}$ are the parameters of the prior distribution of $w$. The mean and variance of the random variable $w$ are given 
by:
\begin{align}
\label{eq:gamma-mean}
\textrm{E}[w] &= \alpha / \beta , \\
\label{eq:gamma-var}
\textrm{var}[w] &= \alpha / \beta^2.
\end{align}

\section{EM with Random Weights} 
\label{sec:em_formulation}
In this section we derive the WD-EM algorithm associated to a model in which the weights are treated as random variables following\eref{eq:w_gamma}. 
The gamma distribution of each $w_i$ is assumed to be parameterized by $\phi_i=\{\alpha_i,\beta_i\}$. Within this framework, the expectation 
of the complete-data log-likelihood is computed over both the assignment and weight variables:
\begin{equation}
 \mathcal{Q}_\textsc{r}\big(\Thetavect,\Thetavect\ir\big) = \E_{P(\Zvect,\Wvect|\Xvect; \Thetavect\ir,\Phivect)}[\ln
P(\Zvect,\Wvect,\Xvect ; \Thetavect,\Phivect)],
\label{eq:q_r_function}
\end{equation}
where we used the notation $\Phivect=\{\phivect_1,\hdots,\phivect_n\}$.
We notice that the posterior distribution factorizes on $i$: 
\[ P\big(\Zvect,\Wvect|\Xvect ; \Thetavect\ir,\Phivect\big)=\prod_{i=1}^n
P\big(z_i,w_i|\xvect_i; \Thetavect\ir,\phivect_i\big)\]
and each one of these factors can be be decomposed as:
\begin{align}
\nonumber P\big(z_i,w_i|\xvect_i ; \Thetavect\ir,\phivect_i\big) & = \\  P(w_i | z_i, \xvect_i ;  \Thetavect\ir, & \phivect_i) 
 P(z_i |  \xvect_i ; \Thetavect\ir, \phivect_i),\label{eq:joint_posterior}
\end{align}
where the two quantities on the right-hand side of this equation have closed-form expressions. The computation of each one of these two expressions leads to two sequential steps, the E-W-step and the E-Z-step, of the expectation step of the proposed algorithm. 

\subsection{The E-Z Step}
The marginal posterior distribution of
$z_i$ is obtained by integrating\eref{eq:joint_posterior} over $w_i$. As previously, we denote the responsibilities with
$ \eta_{ik}\irp =P(z_i = k  |  \xvect_i ; \Thetavect\ir, \phivect_i)$. The integration computes:
\begin{align}
 \eta_{ik}\irp & =  \int P \left(z_i=k ,w_i|\xvect_i ; \Thetavect\ir, \phivect_i \right) d w_i \nonumber \\
& \stackrel{k}{\propto} \int \pi_k\ir \; P\left( \xvect_i | z_i =k ,w_i ; \Thetavect\ir\right)  \;  P(w_i ; \phivect_i) \; d w_i \nonumber \\
& = \int \pi_k\ir \; \hat{p}\left( \xvect_i ; \thetavect_k\ir, w_i\right)  \;  {\cal G}(w_i ; \alpha_i,\beta_i) \; d w_i \nonumber \\
\label{eq:responsibilities}
&\propto \pi_k\ir \;  {\cal P}(\xvect_i; \muvect_k\ir, \Sigmamat_k\ir, \alpha_i, \beta_i),
\end{align}
where 
${\cal P}(\xvect_i; \muvect_k, \Sigmamat_k, \alpha_i, \beta_i)$ denotes the Pearson type VII probability distribution function, which can  be seen as a generalization of the t-distribution:
\begin{align}
\nonumber & {\cal P}(\xvect; \muvect, \Sigmamat, \alpha, \beta) =  \\
\label{eq:pearsonVII}
&\frac{\Gamma(\alpha + d/2)}{|\Sigmamat|^{1/2} \;\Gamma(\alpha) \; (2 \pi \beta)^{d/2}} 
   \left(1+ \frac{ \mah{\xvect-\muvect}{\Sigmamat}^2}{2\beta}\right)_{.}^{-(\alpha+\frac{d}{2})}
\end{align}

\subsection{The E-W Step}
The posterior distribution of $w_i$, namely $p(w_i | z_i =k, \xvect_i; \Thetavect\ir, \phivect_i)$ is a gamma distribution, because it is the conjugate prior of the 
precision of the Gaussian distribution. Therefore, we only need to compute the parameters of the posterior gamma distribution:
\begin{align}
 \nonumber P(w_i | z_i & =k , \xvect_i ;  \Thetavect\ir,\phivect_i) 
 \\ 
 \stackrel{w_i}{\propto} & \; P( \xvect_i | z_i =k ,w_i ; \Thetavect\ir)   P(w_i ; \phivect_i)  \nonumber \\
=&  \; {\cal N}(\xvect_i; \muvect_k\ir, \Sigmamat_k\ir/{w_i}) \; {\cal G}(w_i;  \alpha_i, \beta_i) \nonumber \\
=&   \; {\cal G}(w_i;  a_i\irp, b_{ik}\irp),  \label{newEW}
\end{align}
where the parameters of the posterior gamma distribution are evaluated with:
\begin{align}
\label{eq:alpha-post}
 a_i\irp &= \alpha_i + \frac{d}{2},\\
 \label{eq:beta-post}
 b_{ik}\irp &= \beta_i + \frac{1}{2} \mah{\xvect_i-\muvect_k\ir}{\Sigmamat_k\ir}^2
\end{align}
The conditional mean of $w_i$, namely $\overline{w}_{ik}\irp$, can then be evaluated with:
\begin{equation} 
\label{eq:conditional-w}
\overline{w}_{ik}\irp = \E_{P(w_i | z_i =k, \xvect_i; \Thetavect\ir, \phivect_i)}[w_i] = \frac{a_i\irp }{ b_{ik}\irp }. 
\end{equation}
\addnote[w-effect]{1}{
While estimating the weights themselves is not needed by the algorithm, it is useful to evaluate them in order to fully characterize the 
observations and to discriminate between inliers and outliers. First notice that the marginal  posterior distribution of $w_i$ is a 
mixture of gamma distributions:
\begin{align}
p(w_i | &  \xvect_i;  \Thetavect\ir, \phivect_i)  \nonumber \\
 = &  \sum_{k=1}^K p(w_i | z_i=k, \xvect_i ; \Thetavect\ir,\phivect_i) p(z_i=k | \xvect_i;\Thetavect\ir,\phivect_i ) \nonumber \\
= &\sum_{k=1}^K {\cal G}(w_i;  a_i\irp, b_{ik}\irp)  \;  \eta_{ik}\irp,
\end{align}
and therefore the posterior mean of $w_i$ is evaluated with:
\begin{equation}
\overline{w}_i\irp = \E[w_i | \xvect_i;\Thetavect\ir, \phivect_i] = \sum_{k=i}^K  \eta_{ik}\irp \overline{w}_{ik}\irp.
\end{equation}
By inspection of \eref{eq:alpha-post}, \eref{eq:beta-post}, and \eref{eq:conditional-w} it is easily seen that the value of $\overline{w}_i$ decreases 
as the distance between the cluster centers and observation $\xvect_i$ increases. Importantly, the evaluation of $\overline{w}_i$ enables outlier 
detection. Indeed, an outlier is expected to be far from all the clusters, and therefore all $\overline{w}_{ik}$ will be small, leading to a 
small value of $\overline{w}_i$. It is worth noticing that this is not possible using only the responsibilities $\eta_{ik}$, since they are 
normalized by definition, and therefore their value is not an absolute measure of the datum's relevance, but only a relative measure of it.
}
\subsection{The Maximization Step}

This step maximizes the expected complete-data log-likelihood over the mixture parameters.  By expanding\eref{eq:q_r_function}, we have:
\begin{align}
\nonumber \mathcal{Q}_\textsc{r}\big(\Thetavect, \Thetavect\ir\big)
& \\
\stackrel{\Thetavect}{=} & \; \sum_{i=1}^n\sum_{k=1}^K\int_{w_i} \eta_{ik}\irp\ln \pi_k
\mathcal{N}\Big(\xvect_i;\muvect_k, \frac{1}{w_i}\Sigmamat_k\Big) \nonumber \\
\times & \; p(w_i | \xvect_i, z_i=k, \Thetavect\ir, \phivect_i) \textrm{d} w_i  \nonumber \\
= &\sum_{i=1}^n\sum_{k=1}^K\eta_{ik}\irp \Big( \ln\pi_k - \ln\det{\Sigmamat_k}^{1/2}  \nonumber \\
\label{eq:variable-mstep}
- & \frac{\overline{w}_{ik} \irp}{2}(\xvect_i-\muvect_k)\tr\Sigmamat_k^{-1}(\xvect_i-\muvect_k)\Big).
\end{align}

The parameter updates are obtained from canceling out the derivatives of the expected complete-data
log-likelihood\eref{eq:variable-mstep}. As with standard Gaussian mixtures, all the updates are 
closed-form expressions:
\begin{equation}
\label{eq:var_pi}
 \pi_k\irp =\frac{1}{n}\sum_{i=1}^n\eta_{ik}\irp,
 \end{equation}
 \begin{equation}
\label{eq:var_mean}
 \muvect_k\irp = \frac{\displaystyle\sum_{i=1}^n \overline{w}_{ik}\irp\eta_{ik}\irp\xvect_i}{\displaystyle\sum_{i=1}^n
\overline{w}_{ik}\irp\eta_{ik}\irp},
\end{equation}
\begin{equation}
\label{eq:var_cov}
\Sigmamat_k\irp = \frac{\displaystyle\sum_{i=1}^n \eta_{ik}\irp \overline{w}_{ik}\irp
\left(\xvect_i-\muvect_k\irp\right)\left(\xvect_i-\muvect_k\irp\right)\tr}{\displaystyle\sum_{i=1}^n \eta_{ik}\irp}.
\end{equation}
It is worth noticing that the M-step of the WD-EM algorithm is very similar to the M-step of the FWD-EM algorithm (section \ref{sec:EM-fixed}). 
Indeed, the above iterative formulas, \eref{eq:var_pi}, \eref{eq:var_mean}, \eref{eq:var_cov}  are identical to the formulas \eref{eq:prior-fixed}, 
\eref{eq:mean-fixed}, \eref{eq:cov-fixed}, except that the fixed weights $w_i$ are here replaced with the posterior means of the random weights, 
$\overline{w}_{ik}\irp$.

\section{Estimating the Number of Components}
\label{sec:model-selection}
So far it has been assumed that the number of mixture components $K$ is provided in advance. This assumption is unrealistic for most real-world applications. In this Section we propose to extend  the method and algorithm proposed in \cite{figueiredo2002unsupervised} to the weighted-data clustering model. 
An interesting feature of this model selection method is that it does not require parameter estimation for many different values of $K$, as it would be the case with the Bayesian information criterion (BIC) \cite{schwarz1978estimating}. Instead, the algorithm starts with a large number of components and iteratively deletes components as they become irrelevant. Starting with a large number of components has the additional advantage of making the algorithm robust to initialization.
Formally, the parameter estimation problem is cast into a transmission encoding problem and the criterion is to minimize the expected length of 
the message to be transmitted:
\begin{equation}
 \textrm{length}(\Xvect,\Thetamat) = \textrm{length}(\Thetamat) + \textrm{length}(\Xvect|\Thetamat).
\end{equation}
In this context, the observations and the parameters have to be quantized to finite precision before the transmission. This quantization sets a trade 
off between the two terms of the previous equation. Indeed, when truncating to high precision, 
$\textrm{length}(\Thetamat)$ may be long, but $\textrm{length}(\Xvect|\Thetamat)$ will be short, since the parameters fit well the data. Conversely, 
if the quantization is coarse, $\textrm{length}(\Thetamat)$ may be short, but $\textrm{length}(\Xvect|\Thetamat)$ will be long. The optimal 
quantization step can be found by means of the Taylor approximation~\cite{figueiredo2002unsupervised}. In that case, the optimization 
problem corresponding to the \textit{minimum message length} (MML) criterion, is:
\begin{align}
 \nonumber\Thetamat_{\textrm{MML}} &= \argmin_{\Thetamat} \Big\{-\log \prob(\Thetamat) - \log\prob(\Xvect|\Thetamat,\Phimat) \\
 & + \frac{1}{2}\log\abs{\Imat(\Thetamat)} + \frac{{\cal D}(\Thetamat)}{2}\Big(1 + \log\frac{1}{12}\Big)\Big\},
\label{eqn:mml_single}
\end{align}
where $\Imat(\Thetamat) = -\E\{D^2_{\Thetamat}\log\prob(\Xvect|\Thetamat)\}$ is 
the \textit{expected} Fisher information matrix (FIM) and ${\cal D}(\Thetamat)$ denotes the dimensionality of the model, namely the dimension of the parameter vector $\Thetamat$. Since the minimization \eref{eqn:mml_single} does not depend on the weight parameters, $\Phimat$ will be omitted for simplicity.

In our particular case, as in the general case of mixtures, the Fisher information matrix cannot be obtained analytically. Indeed, the direct 
optimization of the log-likelihood does not lead to closed-form solutions. Nevertheless, it was noticed that the
\textit{complete} FIM upper bounds the FIM~\cite{figueiredo2002unsupervised}, and that the expected complete-data 
log-likelihood lower bounds the log-likelihood. This allows us to write the following equivalent optimization problem:
\begin{align}
 \nonumber\Thetamat_{\textrm{MML}} &= \argmin_{\Thetamat} \Big\{-\log \prob(\Thetamat) - \log\mathcal{Q}_\textsc{r}\big(\Thetavect, 
\Thetavect\ir\big) \\
 & + \frac{1}{2}\log\abs{\Imat_c(\Thetamat)} + \frac{{\cal D}(\Thetamat)}{2} \Big(1 + \log\frac{1}{12}\Big) \Big\},
\label{eqn:mml_mixt}
\end{align}
where $\Imat_c$ denotes the expected complete-FIM and ${\cal Q}_\textsc{r}$ is evaluated with~\eref{eq:variable-mstep}.

As already mentioned, because there is a different weight $w_i$ for each observation $i$, the observed data are not identically distributed 
and our model cannot be considered a classical mixture model. For this reason, the algorithm proposed in \cite{figueiredo2002unsupervised} cannot be applied directly to our model. Indeed, in 
the proposed WD-GMM setting, the complete-FIM is:
\begin{equation}
 \Imat_c(\Thetamat) \!=\! \mbox{diag}\Big(\pi_1\!\sum_{i=1}^n \Imat_i(\thetavect_1),\ldots,\pi_K\!\sum_{i=1}^n \Imat_i(\thetavect_K),n\Mmat 
\Big) \!\! \label{eq:complete-fim}
\end{equation}
where $\Imat_i(\thetavect_k)= -\E\{D^2_{\thetavect_k}\log{\cal P}(\xvect_i |\thetavect_k,\alpha_i,\beta_i)\}$ is the Fisher information 
matrix for the $i$-th observation with respect to the parameter vector $\thetavect_k$ (mean and the covariance) of the $k$-th component, ${\cal P}$ is defined 
in~(\ref{eq:pearsonVII}), and $\Mmat$  is the Fisher information matrix of the multinomial distribution, namely the diagonal matrix $\mbox{diag}(\pi_1^{-1},\ldots,\pi_K^{-1})$. We can evaluate $\abs{\Imat_c(\Thetamat)}$ from~\eref{eq:complete-fim}:
\begin{equation}
 \abs{\Imat_c(\Thetamat)} = n^{K(M+1)}|\Mmat|\prod_{k=1}^K\pi_k^M\left|\frac{1}{n}\sum_{i=1}^n\Imat_i(\thetavect_k)\right|,\label{eq:det-fim}
\end{equation}
where $M$ denotes the number of free parameters of each component. For example, $M=2d$ when using diagonal covariance matrices or $M=d(d+3)/2$ when 
using full covariance matrices.

Importantly, one of the main advantages of the methodology proposed in \cite{figueiredo2002unsupervised} is that one has complete freedom to choose 
a prior distribution on the parameters, $\prob(\Thetamat)$. In our case, inspired by~(\ref{eq:det-fim}), we select the following prior 
distributions for the parameters:
\begin{align}
 &\prob(\thetavect_k) \propto \left|\frac{1}{n}\sum_{i=1}^n \Imat_i(\thetavect_k)\right|^{-\frac{1}{2}}, \label{eq:prior-component}\\
 &\prob(\pi_1,\ldots,\pi_K) \propto |\Mmat|^{-\frac{1}{2}}. \label{eq:prior-proportions}
\end{align}
By substitution of (\ref{eq:det-fim})--(\ref{eq:prior-proportions}) into~(\ref{eqn:mml_mixt}) we obtain the following optimization problem:
\begin{align}
 \nonumber\Thetamat_{\textrm{MML}} = \argmin_{\Thetamat} & \Big\{\frac{M}{2}\sum_{k=1}^K\log\pi_k - \log\mathcal{Q}_\textsc{r}\left(\Thetavect, 
\Thetavect\ir\right) \\
 & + \frac{K(M+1)}{2}\left(1 + \log\frac{n}{12}\right)\Big\},
\label{eqn:mml_final}
\end{align}
where we used ${\cal D}(\Thetamat) = K(M+1)$.

One may notice that (\ref{eqn:mml_final}) does not make sense (diverges) if any of the $\pi_k$'s is allowed to be null. However, in the current length coding
framework, there is no point in transmitting the parameters of an empty component. Therefore, we only focus on the non-empty components, namely 
those components for which $\pi_k>0$. Let ${\cal K}^+$ denote the index set of non-empty components and let $K^+=|{\cal K}^+|$ be its cardinality. We can rewrite 
(\ref{eqn:mml_final}) as:
\begin{align}
 \nonumber\Thetamat_{\textrm{MML}} = \argmin_{\Thetamat} & \Big\{\frac{M}{2}\sum_{k\in{\cal K}^+}\log\pi_k - 
\log\mathcal{Q}_\textsc{r}\left(\Thetavect, \Thetavect\ir\right) \\
 & + \frac{K^+(M+1)}{2}\left(1 + \log\frac{n}{12}\right)\Big\}.
\label{eqn:mml_final_nz}
\end{align}

The above minimization problem can be solved by modifying the EM algorithm described in \sref{sec:em_formulation} (notice that there is an equivalent derivation for the fixed-weigth EM algorithm described in \sref{sec:EM-fixed}). Indeed, we remark that the minimization \eref{eqn:mml_final_nz} is equivalent to using a symmetric improper Dirichlet prior for the proportions with exponent $-M/2$. Moreover, since the 
optimization function for the parameters of the Gaussian components is the same (equivalently, we used a flat prior for the mean vector and 
covariance matrix), their estimation formulas \eqref{eq:var_mean} and \eqref{eq:var_cov} still hold. Therefore, we only need to modify the estimation of the mixture proportions, namely:
\begin{equation}
 \pi_k = \frac{\max \left\{0,\sum_{i=1}^{n}\eta_{ik}-\frac{M}{2}\right\}}{\sum_{k'=1}^{K}\max 
\left\{0,\sum_{i=1}^{n}\eta_{ik'}-\frac{M}{2}\right\}}. 
\label{eqn:pi_update_mml}
\end{equation}
The $\max$ operator in \eqref{eqn:pi_update_mml} verifies whether the $k$-th component is supported by the data. When one of the components 
becomes too weak, \ie the required minimum support $M/2$ cannot be obtained from the data, this component is annihilated. In other words, 
its parameters will not be estimated, since there is no need in transmitting them. One has to be careful in this context, since starting with a 
large value of $K$ may lead to several empty components. In order to avoid this singular situation, we adopt the component-wise EM procedure (CEM) \cite{celeux2001component}, 
as proposed in \cite{figueiredo2002unsupervised} as well. Intuitively, we run both E and M steps for one component, before moving to the next 
component. More precisely, after running the E-Z and E-W steps for the component $k$, its parameters are updated if $k\in{\cal K}^+$, otherwise the 
component is annihilated if $k\not\in{\cal K}^+$. The rationale behind this procedure is that, when a component is annihilated its probability mass 
is immediately redistributed among the remaining components. Summarizing, CEM updates the components one by one, whereas the classical 
EM simultaneously updates all the components. 

The proposed algorithm is outlined in \refalg{alg:cem_for_wdgmm}. In practice, an upper and a lower number of components, $K_{\up}$ and $K_{\low}$, 
are provided. Each iteration $r$ of the algorithm consists of component-wise E and M steps. If needed, some of the components are annihilated, and the 
parameters are updated accordingly, until the relative length difference is below a threshold, $\abs{\Delta\mathbf{LEN}_\text{MML}^{(r)}} < 
\varepsilon$. In that case, if the message length, \ie~(\ref{eqn:mml_final_nz}) is lower than the current optimum, the parameters, weights, and 
length are saved in $\Thetamat_{\min}$, $W_{\min}$ and $\mathbf{LEN}_{\min}$ respectively. In order to explore the full range of $K$, the less 
populated component is artificially annihilated, and CEM is run again. \addnote[complexity]{3}{The complexity of \refalg{alg:cem_for_wdgmm} is similar to the complexity 
of the algorithm in~\cite{figueiredo2002unsupervised}, with the exception of the E-W step. However, the most computationally intensive part of this step (matrix inversion and matrix-vector multiplications in \eqref{eq:beta-post}) is already achieved in the E-Z step.}

\begin{figure}[!t]
\resizebox{.43\textwidth}{!}{
\removelatexerror
\begin{algorithm}[H]
\DontPrintSemicolon
\SetKwData{Left}{left}\SetKwData{This}{this}\SetKwData{Up}{up}
\SetKwFunction{Union}{Union}\SetKwFunction{FindCompress}{FindCompress}
\SetKwInOut{Input}{input}\SetKwInOut{Output}{output}
\Input{$\Xvect=\{\xvect_i\}_{i=1}^n, K_{\low}, K_{\up}, \Thetamat^{(0)}=\{\pi_k^{(0)},\muvect_k^{(0)},\Sigmamat^{(0)}_k\}_{K=1}^{K_{\up}}, 
\Phimat^{(0)}=\{\alpha_i^{(0)},\beta_i^{(0)}\}_{i=1}^n$}
\Output{The minimum length mixture model: $\Thetamat_{\min}$ and the final data weights: $\Wvect_{\min}$}
\BlankLine
 Set: $r=0, {\cal K}^+=\{k\}_{k=1}^{K_{\up}}, \mathbf{LEN}_{\min} = +\infty$\; 
    \While{$|{\cal K}^+|\geq K_{\low}$}{
      \Repeat{$\abs{\Delta\mathbf{LEN}_\text{MML}^{(r)}} < \varepsilon $}{
	\For{$k=1$ \KwTo $K_{\up}$}{
	  E-Z step using~(\ref{eq:responsibilities}):\;
	    $\qquad \eta_{ik}^{(r+1)} = 
    \frac{\pi^{(r)}_k{\cal P}\left(\xvect_i;\muvect_k^{(r)},\Sigmamat_k^{(r)},\alpha_i^{(r)},\beta_{ik}^{(r)}\right)} 
    {\sum_{l=1}^{K_{\up}}\pi^{(r)}_l{\cal P}\left(\xvect_i;\muvect_l^{(r)},\Sigmamat_l^{(r)},\alpha_i^{(r)},\beta_{il}^{(r)}\right)}$\;
	  \BlankLine
	  E-W step using~(\ref{eq:alpha-post})--(\ref{eq:beta-post}):\;
	    $\qquad \alpha_i^{(r+1)} = \alpha_i^{(0)} + \tfrac{d}{2}$ \;
	    $\qquad \beta_{ik}^{(r+1)} = \beta_i^{(0)} + \frac{1}{2} \mah{\xvect_i-\muvect_k^{(r)}}{\Sigmamat_k^{(r)}}^2$ \;
	    $\qquad \overline{w}_{ik} =\frac{\alpha_i^{(r+1)}}{\beta_{ik}^{(r+1)}}$ \;
	  \BlankLine
	  M-step \;
	  $\qquad  \pi_k^{(r+1)} = \frac{\max \left\{0,\sum_{i=1}^{n}\eta_{ik}^{(r+1)}-\frac{M}{2}\right\}}{\sum_{l=1}^{K_{\up}}\max 
    \left\{0,\sum_{i=1}^{n}\eta_{il}^{(r+1)}-\frac{M}{2}\right\}}$ \;
	 \eIf{$\pi_k^{(r+1)}>0$}{
	  Evaluate $\thetavect_k^{(r+1)}$: mean $\muvect_k^{(r+1)}$ using \eref{eq:var_mean} and covariance $\Sigmamat_k^{(r+1)}$ using \eref{eq:var_cov}.
	  }{
	    $K^+ = K^+-1$\;
	  }
	} 
	$\Thetamat^{(r+1)} = \left\{\pi_k^{(r+1)},\thetavect_k^{(r+1)}\right\}_{k=1}^{K_{\up}}$ \;
	Compute optimal length $\mathbf{LEN}_\text{MML}^{(r+1)}$ with (\ref{eqn:mml_final_nz}).\;
	$r\leftarrow r+1$\;
	\BlankLine
      }
      \BlankLine
      \If{$\mathbf{LEN}_\text{MML}^{(r)} < \mathbf{LEN}_{\min}$}{
        $\mathbf{LEN}_{\min} = \mathbf{LEN}_\text{MML}^{(r)}$\;
	$\Thetamat_{\min} = \Thetamat^{(r)} $ \;
	$\Wvect_{\min}=\{\overline{w}_i\}_{i=1}^n$ with $\overline{w}_i = \sum\limits_{k=1}^{K_{\up}}\eta_{ik}\overline{w}_{ik}$ \;
      }
      $k^* = \argmin_{k'\in{\cal K}^+}\left(\pi_{k'}^{(r)}\right), \quad {\cal K}^+ = {\cal K}^+/ k^* $
    }
\caption{WD-EM with model selection based on the MML criterion.
}
\label{alg:cem_for_wdgmm}
\end{algorithm}
}
\end{figure}

\section{Algorithm Initialization}
\label{sec:initialization}
\addnote[init-new]{1}
{
The EM algorithms proposed in \sref{sec:EM-fixed}, \sref{sec:em_formulation}, and \sref{sec:model-selection} require proper initialization of both the weights (one for each observation and either a fixed value $w_i$ or parameters $\alpha_i, \beta_i$) and of the model parameters. The $K$-means 
algorithm is used for an initial clustering, from which values for the model parameters are computed. In this section we concentrate onto the issue of weight initialization. An interesting feature 
of our method is that the only constraint on the weights is that they must be positive. 
Initial $w_i$ values may depend 
on expert or prior knowledge and may be experiment- or goal-dependent. This model flexibility allows the incorporation of such prior 
knowledge. 
In the absence of any prior information/knowledge, we propose  a data-driven initialization scheme
and make the assumption that densely sampled 
regions are more important that sparsely sampled ones. We note that a similar strategy could be used if one wants to reduce the importance of dense data and to give more importance to small groups of data or to sparse data.
}

We adopt a well known data similarity measure based on the Gaussian kernel, and it follows that the weight $w_i$ associated with the 
data point $i$ is evaluated with:
\begin{equation}
\label{eqn:weightformula}
 w_i = \sum\limits_{j\in{\cal S}_i^q} \exp\left(-\dfrac{d^2(\xvect_i,\xvect_j)}{\sigma}\right),
\end{equation}
where $d(\xvect_i,\xvect_j)$ is the Euclidean distance, ${\cal S}_i^q$ denotes the set containing the $q$ nearest neighbors
of $\xvect_i$, and $\sigma$ is a positive scalar. In all the experiments we used $q=20$ for the simulated datasets and $q=50$ for the real datasets. In both cases, we used $\sigma=100$.
In the case of the FWD-EM algorithm, the weights $w_i$ thus initialized remain unchanged. However, in the case of the WD-EM algorithm, the weights are modeled as latent random variables drawn from a gamma 
distribution, hence one needs to set  initial values for the parameters of this
distribution, namely $\alpha_i$ and $\beta_i$ in~\eref{eq:w_gamma}. Using~\eref{eq:gamma-mean} and~\eref{eq:gamma-var} one can choose to initialize these parameters such as
$\alpha_i = w_i^2 $ and $\beta_i =  w_i$, such that the mean and variance of the prior distribution are $ 
w_i$ and $1$ respectively. 

\section{Experimental Validation}
\label{sec:experiments}

\begin{figure*}[hbt]
\begin{minipage}[b]{2\columnwidth}
\centering
\subfloat{\includegraphics[trim=7mm 4mm 2mm 7mm,clip,scale=0.25]{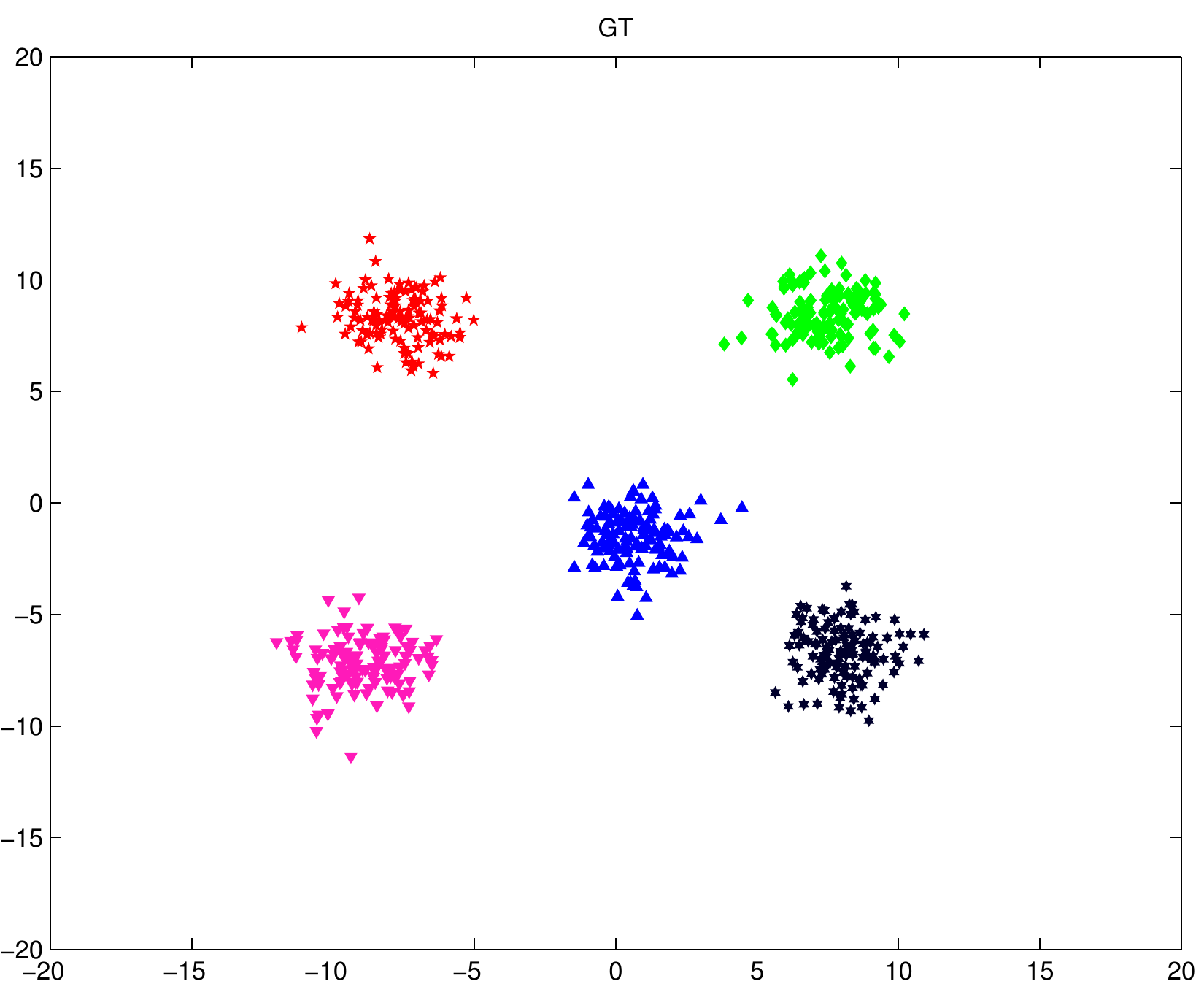}} \
\subfloat{\includegraphics[trim=7mm 4mm 2mm 7mm,clip,scale=0.25]{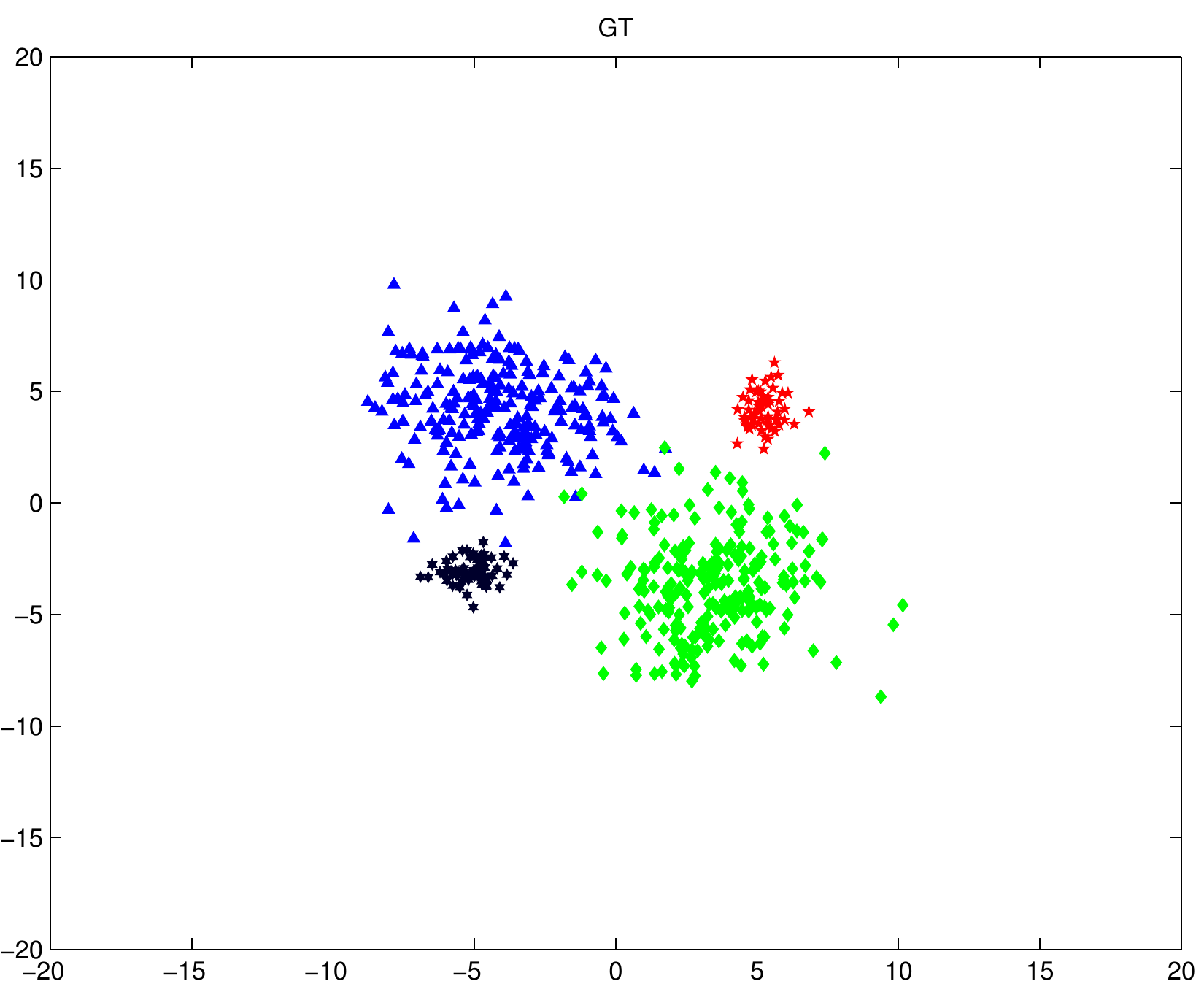}} \
\subfloat{\includegraphics[trim=7mm 4mm 2mm 7mm,clip,scale=0.25]{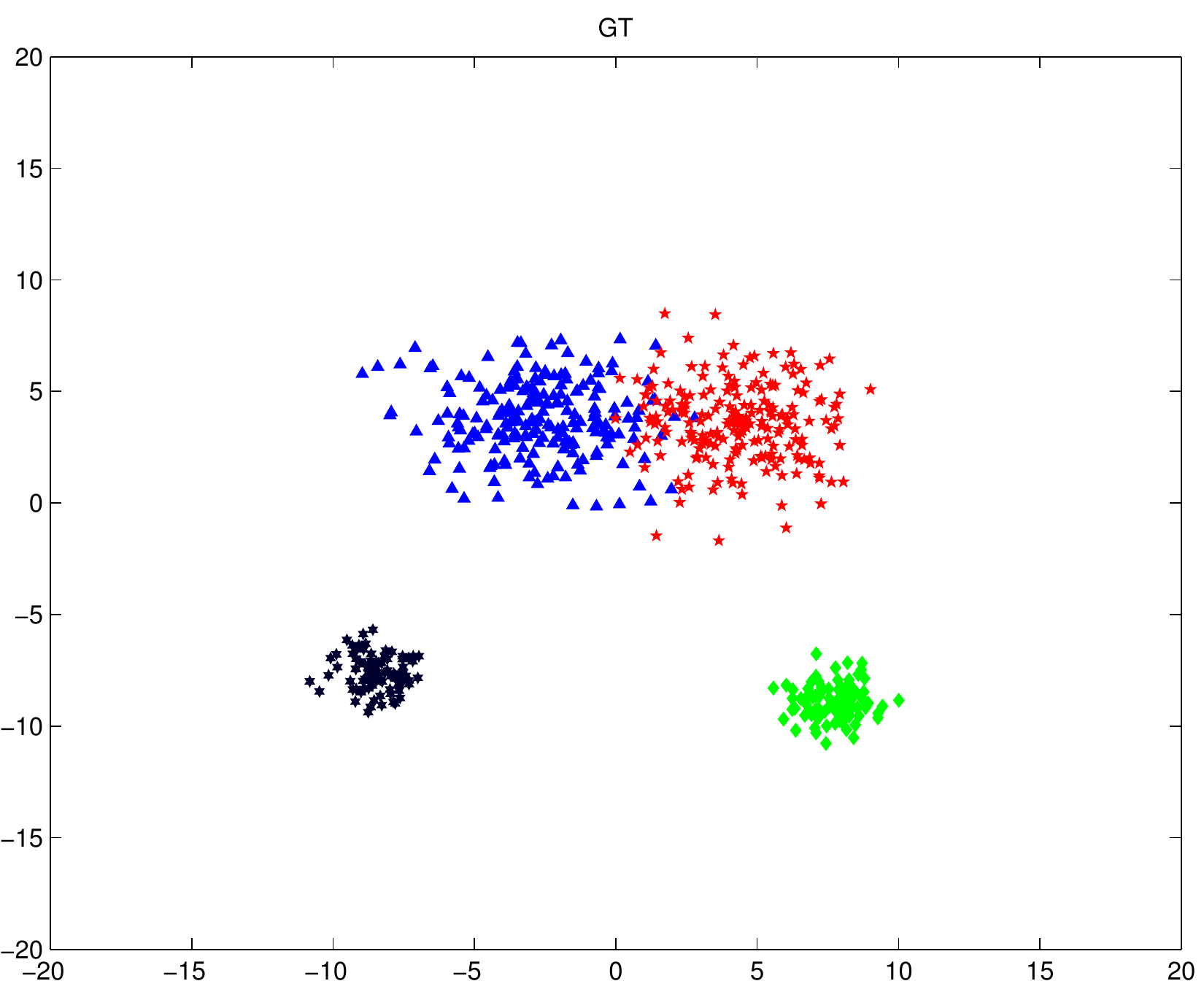}}\
\subfloat{\includegraphics[trim=7mm 4mm 2mm 7mm,clip,scale=0.25]{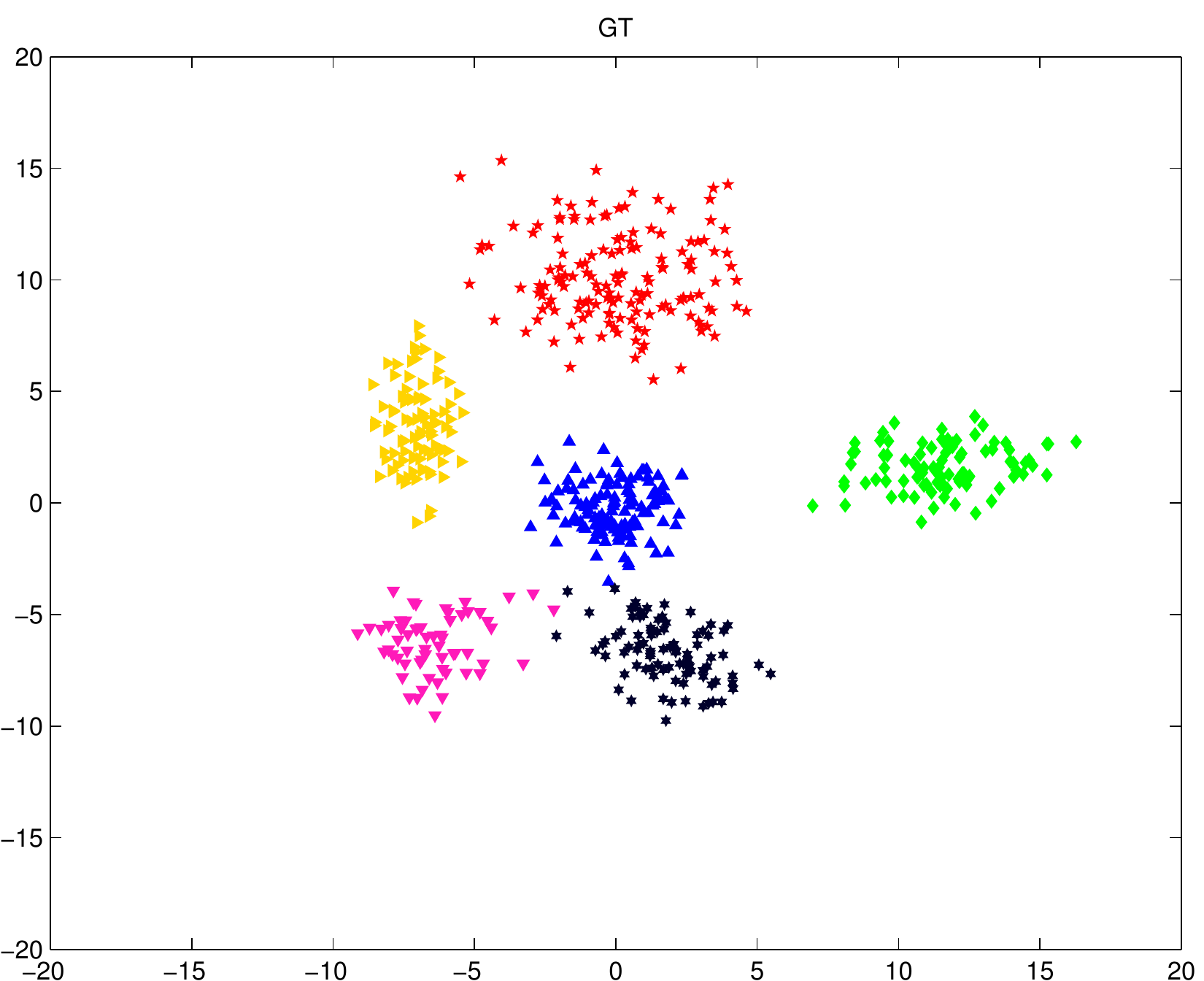}}\
\end{minipage}
\begin{minipage}[b]{2\columnwidth}
\setcounter{subfigure}{0}
\centering
\subfloat[SIM-Easy ]{\includegraphics[trim=7mm 4mm 2mm 7mm,clip,scale=0.25]{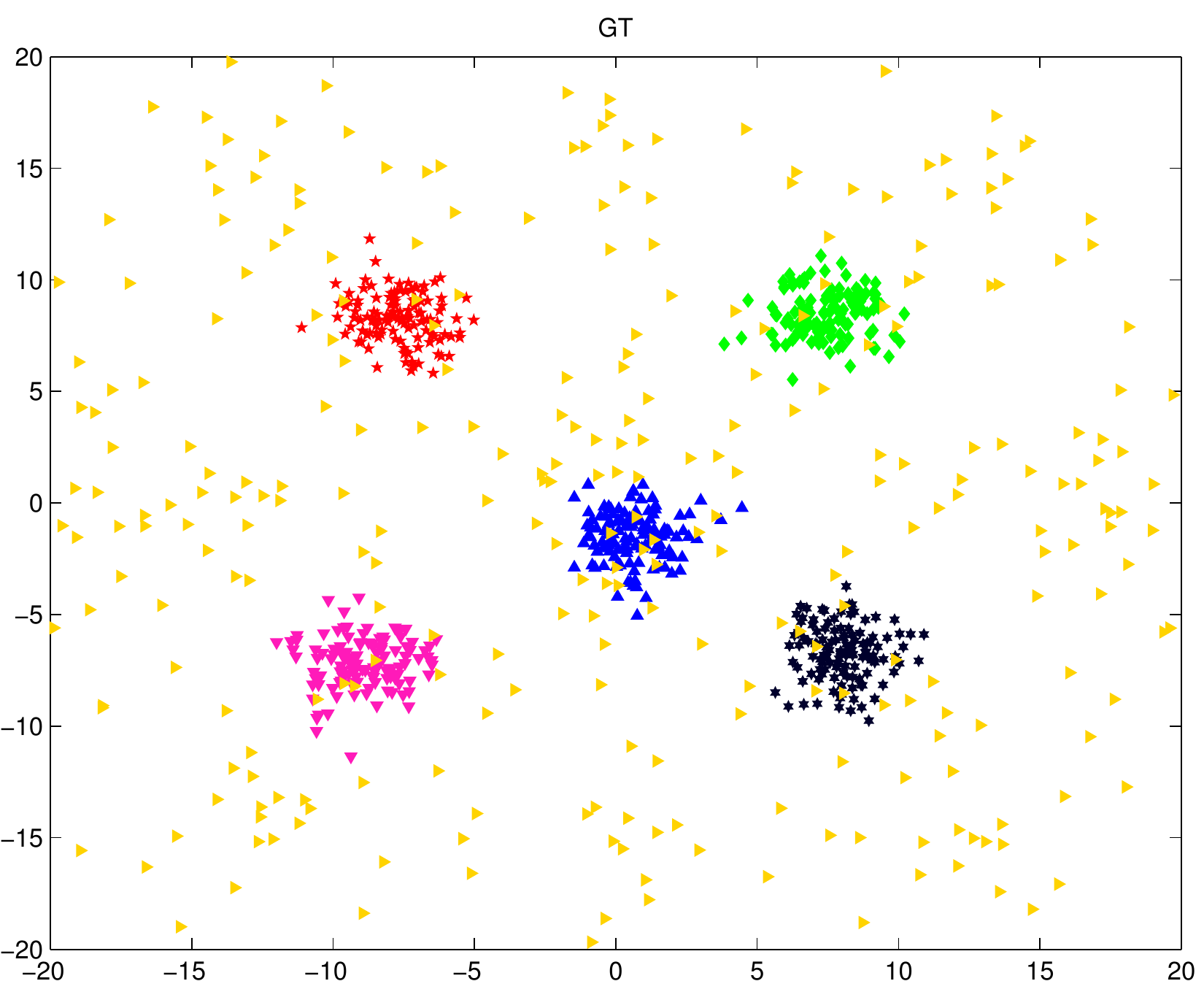}} \
\subfloat[SIM-Unbalanced]{\includegraphics[trim=7mm 4mm 2mm 7mm,clip,scale=0.25]{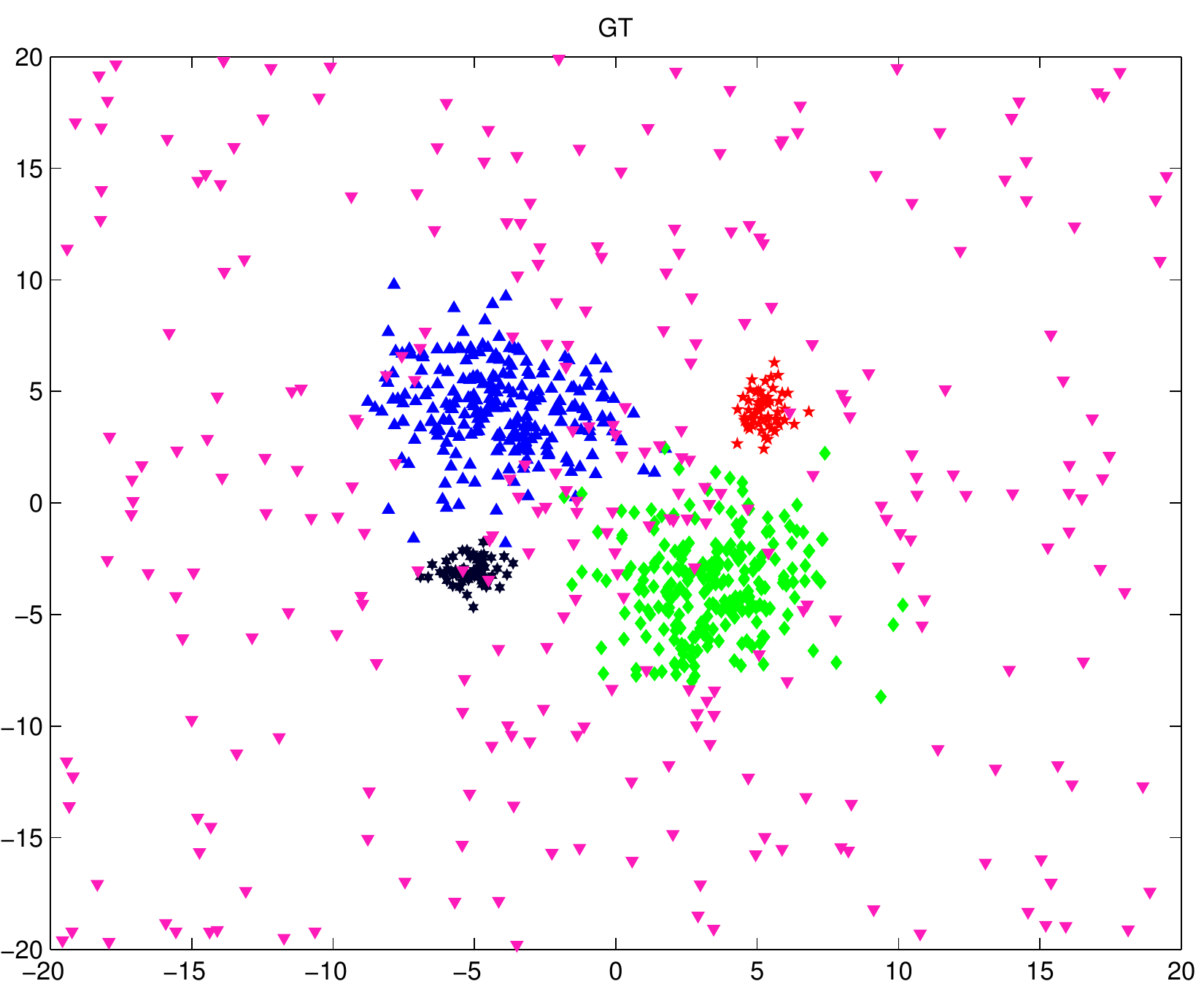}} \
\subfloat[SIM-Overlapped]{\includegraphics[trim=7mm 4mm 2mm 7mm,clip,scale=0.25]{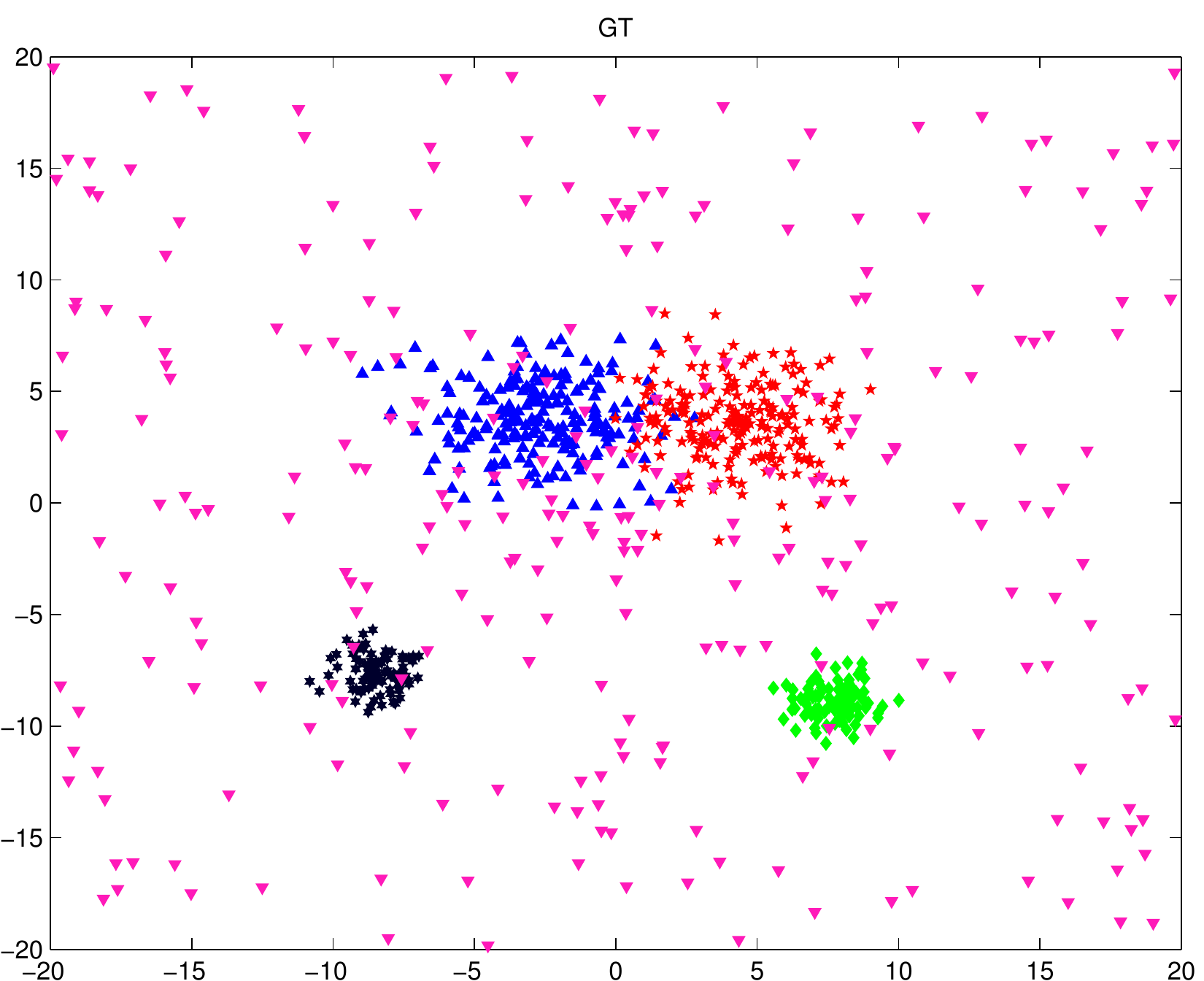}}\
\subfloat[SIM-Mixed]{\includegraphics[trim=7mm 4mm 2mm 7mm,clip,scale=0.25]{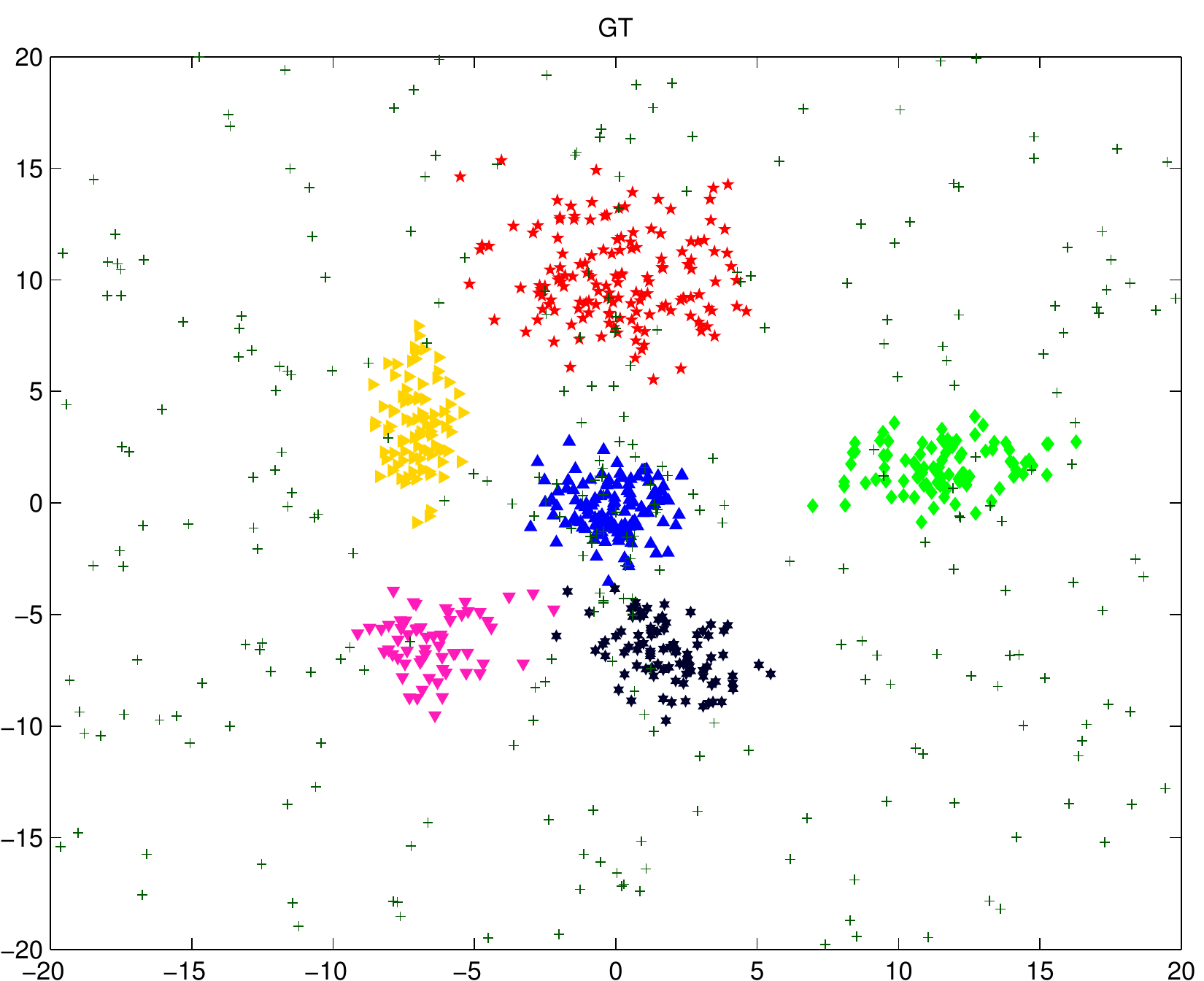}}\
\caption{\label{fig:sim_dataset} Samples of the SIM dataset with no outliers (top row) and contaminated with 50\% outliers (bottom row). The 600 inliers are generated from Gaussian mixtures while the 300 outliers are generated from a uniform distribution.}
\end{minipage}
\end{figure*} 
\begin{table}[b]
\centering
\caption{Datasets used for benchmarking and their characteristics: $n$ is the number of data points, $d$ is the dimension of the data space, and $K$ is number of clusters.}
\begin{tabular}{lrrr}
\toprule
Data Set & $n$ & $d$ & $K$ \\
\midrule
SIM-Easy & $600$  & $2$   & $5$  \\
SIM-Unbalanced & $600$  & $2$   & $4$  \\
SIM-Overlapped & $600$  & $2$   & $4$  \\
SIM-Mixed & $600$  & $2$   & $6$  \\
MNIST \cite{lecun1998gradient} & $10,000$   & $141$ & $10$ \\ 
Wav \cite{breiman1984classification}& $5,000$    & $21$  & $3$  \\
BCW \cite{street1993nuclear} & $569$ & $30$  & $2$  \\
Letter Recognition \cite{frey1991letter} & $20,000$ &$16$ & $26$\\
\bottomrule
\end{tabular} %
\label{tab:dataset}
\end{table}

The proposed algorithms were tested and evaluated using eight datasets: four simulated datasets and four publicly available datasets that are widely used for benchmarking clustering methods. The main 
characteristics of these datasets are summarized in \tref{tab:dataset}.
The simulated datasets (SIM) are designed to evaluate the robustness of the proposed method with respect to outliers.  The simulated inliers are drawn from Gaussian mixtures while the simulated outliers 
are drawn from a uniform distribution, \eg \fref{fig:sim_dataset}. The SIM datasets have different cluster configurations in terms of separability, shape and compactness. The eight datasets that we used are the following:
\begin{itemize}
\item \textbf{SIM-Easy:} Five clusters that are well separated and compact. 
\item \textbf{SIM-Unbalanced:} Four clusters of different size and density.
\item \textbf{SIM-Overlapped:} Four clusters, two of them overlap. 
\item \textbf{SIM-Mixed:} Six clusters of different size, compactness and shape. 
 \item \textbf{MNIST} contains instances of handwritten digit images normalized to the same size \cite{lecun1998gradient}. We preprocessed these data with PCA to reduce 
the dimension from 784 to 141, by keeping $95\%$ of the variance.
 \item \textbf{Wav} is the Waveform Database Generator \cite{breiman1984classification}.
 \item \textbf{BCW} refers to the Breast Cancer Wisconsin data set \cite{street1993nuclear}, in which each instance represents a digitized image of a fine needle aspirate (FNA) of breast mass.
\item \addnote[letter_recognition]{1}{\textbf{Letter Recognition} contains $20,000$ single-letter images that were generated by randomly distorting the images of the 26 uppercase letters from 20 different commercial fonts \cite{frey1991letter}. Each letter/image is described by 16 features. This dataset is available through the UCI machine learning repository.}
\end{itemize}

\begin{table*}[htb]
\caption{Results obtained with the MNIST, WAV, BCW, and Letter Recognition datasets. The clustering scores correspond to the Davies-Bouldin (DB) index. The best results are shown in underlined \underline{\textbf{bold}}, and the second best results are shown in \textbf{bold}. The proposed method yields the best results for the WAV and BCW datasets, while I$^2$GMM yields the best results for the MNIST dataset. Interestingly, the non-parametric methods (K-means, HAC and Ncut) yield excellent results for Letter Recognition.}
\centering
\def\arraystretch{1.5}
\resizebox{\textwidth}{!}{
\begin{tabular}{lccccccccccc}
\toprule
   Dataset 		& WD-EM 	& FWD-EM	   & GMM	   & GMM+U	   & FM-uMST	  & IGMM	   & $\text{I}^2$GMM	& 
K-Means	& KK-Means	& Ncut		& HAC \\ \midrule
   MNIST   		&$2.965(0.15)$& $3.104 (0.21)$& $3.291 (0.14)$& $3.245 (0.09)$ & \bestb{2.443 (0.00)}& $3.555 (0.06)$& \besta{2.430 (0.14)} & $2.986 (0.01)$& $2.980 (0.02)$& $4.760 (0.08)$& $3.178 (0.00)$\\	
   WAV  		&\besta{0.975 (0.00)}& $1.019 (0.00)$& $1.448 (0.03)$& $1.026 (0.04)$& $1.094 (0.10)$& $1.028 (0.02)$& $2.537 (0.35)$	& $1.020 (0.00)$& \bestb{0.975 (0.05)}& $2.781 (0.06)$& $1.089 (0.00)$\\	
   BCW  		&\besta{0.622 (0.00)}& $0.687 (0.00)$& $0.714 (0.00)$& $0.689 (0.00)$& $0.727 (0.00)$& $0.719 (0.00)$& $0.736 (0.09)$	& $0.659 (0.00)$& \bestb{0.655 (0.00)}& $0.838 (0.00)$& $0.685 (0.00)$\\
   Letter Recognition   &$1.690 (0.00)$& $1.767 (0.01)$& $2.064 (0.06)$& $2.064 (0.06)$& $1.837 (0.00)$& $2.341 (0.11)$& $1.724 (0.03)$	& \besta{1.450 (0.02)}& $1.504 (0.03)$& \bestb{1.626 (0.00)}& \bestb{1.626 (0.00)}\\	
\bottomrule
\end{tabular} }
\label{tab:db_realdataset}
\end{table*}

\begin{table*}[htb!]
\caption{Micro $\text{F}_1$ scores obtained on the real data sets (MNIST, WAV, BCW and Letter Recognition). The number in parenthesis indicates the standard deviation 
of 20 repetitions. Based on this classification score, $\text{I}^2$GMM yields the best result.}
\centering
\def\arraystretch{1.5}
\resizebox{\textwidth}{!}{
\begin{tabular}{lccccccccccc}
\toprule
Data set		& WD-EM		& FWD-EM	& GMM		& GMM+U		& FM-uMST	& IGMM		& $\text{I}^2$GMM& K-Means	
& KK-Means	& Ncut		& HAC \\ \midrule
MNIST			& $0.524 (0.01)$& $0.455 (0.01)$& \bestb{0.573 (0.00)}& $0.549 (0.01)$ & $0.519 (0.00)$& \besta{0.689 (0.02)} & $0.545 
(0.06)$& $0.497 (0.02)$& $0.507 (0.02)$& $0.402 (0.00)$& $0.532 (0.00)$\\
WAV			& \besta{0.774 (0.00)}& $0.534 (0.00)$&$0.535 (0.00)$& $0.552 (0.00)$ & \bestb{0.632 (0.08)}& $0.543 (0.01)$ & $0.493 (0.00)$& 
$0.521 (0.00)$& $0.522 (0.00)$& $0.387 (0.00)$& $0.597 (0.00)$ \\  
BCW			& \besta{0.965 (0.00)}& $0.907 (0.00)$& $0.885 (0.00)$& $0.915 (0.00)$ & \bestb{0.927 (0.00)}& $0.914 (0.00)$ & $0.682 
(0.00)$& $0.907 (0.00)$& $0.910 (0.00)$& $0.859 (0.00)$& $0.879 (0.00)$\\            
Letter Recognition 	& $0.315 (0.01)$& $0.323 (0.00)$& \bestb{0.423 (0.00)}& \bestb{0.423 (0.00)} & $0.379 (0.00)$& $0.306 (0.02)$ & \besta{0.466 
(0.01)}& $0.340 (0.00)$& $0.343 (0.01)$ &$0.347 (0.00)$& $0.347 (0.00)$\\  
\bottomrule
\end{tabular} }
\label{tab:f1_realdataset}
\end{table*}

In addition to the two proposed methods (FWD-EM and WD-EM) we tested the following algorithms:
\begin{itemize}
 \item \textbf{GMM} uses EM with the standard Gaussian mixture model, implemented as described in \cite{bishop2006pattern};
 \item \textbf{GMM+U} uses EM with  a GMM and with an additional uniform component, \cite{BanfieldRaftery1993};
 \item \textbf{FM-uMST} stands for the \textit{finite mixture of unrestricted multivariate skew t-distribution} algorithm of \cite{lee2014finite};
\item \addnote[igmm]{1}{\textbf{IGMM} stands for the \textit{infinite Gaussian mixture model} \cite{rasmussen1999infinite};}
 \item \addnote[i2gmm]{1}{\textbf{I$^2$GMM} stands for the \textit{infinite mixture of infinite Gaussian mixtures} \cite{yerebakan2014infinite};}
 \item \textbf{$K$-Means} is the standard $K$-means algorithm;
 \item \textbf{K$K$-Means} is the kernel $K$-means algorithm of \cite{dhillon2004kernel};
 \item \textbf{NCUT} is the spectral clustering algorithm of \cite{shi2000normalized}.
 \item \textbf{HAC} is the hierarchical agglomerative clustering algorithm of \cite{zhao2002evaluation}.
\end{itemize}
All the above algorithms need proper initialization. 
All the mixture-based algorithms, WD-EM, FWD-EM, GMM, GMM+U, FM-uMST, IGMM  and $\text{I}^2$GMM start from the same proportions, means, and covariances which are estimated from the set of clusters provided by K-means. The latter is randomly initialized several times to find a good initialization. 
Furthermore, algorithms WD-EM, FWD-EM, GMM, GMM+U and FM-uMST are iterated until convergence, i.e, the log-likelihood difference between two consecutive iterations is less than 1\%, or are stopped after 400 iterations.

\begin{table*}[htb!]
\caption{DB scores obtained on the SIM-X dataset (\underline{\textbf{best}} and \textbf{second best}).}
\centering
\def\arraystretch{1.5}
\resizebox{\textwidth}{!}{
\begin{tabular}{lcccccccccccc}
\toprule
& Outliers	& WD-EM		& FWD-EM	& GMM		& GMM+U		& FM-uMST	& IGMM		& $\text{I}^2$GMM& K-Means	& 
KK-Means	& Ncut		& HAC \\ \midrule
\multirow{5}{*}{\rotatebox[origin=c]{90}{SIM-Easy}}&	$10\%$	& \bestb{0.229 (0.01)}& $0.295 (0.01)$& $0.295 (0.01)$& \besta{0.222 (0.02)}& $0.307 
(0.02)$& $1.974 (0.12)$&$0.500 (0.16)$	 & $0.291 (0.01)$& $0.330 (0.07)$& $0.283 (0.01)$& $0.266 (0.00)$\\
&	$20\%$	& \bestb{0.266 (0.02)}& $0.338 (0.01)$& $0.342 (0.01)$& \besta{0.233 (0.01)}& $0.349 (0.02)$& $1.564 (0.43)$&$0.626 (0.28)$	 & 
$0.344 (0.01)$& $0.420 (0.10)$& $0.335 (0.01)$& $0.330 (0.01)$\\
&	$30\%$	& \bestb{0.330 (0.01)}& $0.385 (0.01)$& $0.384 (0.02)$& \besta{0.227 (0.02)}& $0.501 (0.04)$& $1.296 (0.12)$&$0.570 (0.27)$	 & 
$0.372 (0.01)$& $0.381 (0.03)$& $0.366 (0.02)$& $0.376 (0.01)$\\
&	$40\%$	& \bestb{0.358 (0.01)}& $0.445 (0.04)$& $0.453 (0.05)$& \besta{0.211 (0.02)}& $0.585 (0.06)$& $1.259 (0.16)$&$0.534 (0.21)$	 & 
$0.417 (0.01)$& $0.411 (0.01)$& $0.409 (0.01)$& $0.401 (0.01)$\\
&	$50\%$	& \bestb{0.380 (0.01)}& $0.455 (0.02)$& $0.459 (0.02)$& \besta{0.195 (0.01)}& $0.568 (0.05)$& $1.107 (0.06)$&$0.626 (0.21)$	 & 
$0.422 (0.01)$& $0.439 (0.03)$& $0.422 (0.01)$& $0.438 (0.01)$\\ \midrule
\multirow{5}{*}{\rotatebox[origin=c]{90}{SIM-Unbalanced}}&$10\%$	& \besta{0.270 (0.01)}& $0.954 (0.72)$& $1.354 (1.02)$& \bestb{0.277 (0.01)}& 
$1.104 (0.76)$& $1.844 (0.29)$&$0.491 (0.17)$  & $0.405 (0.02)$& $0.433 (0.05)$& $0.402 (0.02)$& $0.427 (0.02)$\\
	&$20\%$	& \bestb{0.329 (0.03)}& $4.503 (4.33)$& $3.003 (1.85)$& \besta{0.269 (0.01)}& $1.181 (0.44)$& $1.278 (0.45)$&$0.591 (0.13)$  & $0.512 
(0.02)$& $0.515 (0.03)$& $0.477 (0.03)$& $0.529 (0.02)$\\
	&$30\%$	& \bestb{0.399 (0.03)}& $3.502 (3.09)$& $2.034 (1.22)$& \besta{0.252 (0.03)}& $1.414 (0.88)$& $1.272 (0.35)$&$0.601 (0.10)$  & $0.548 
(0.03)$& $0.540 (0.03)$& $0.531 (0.02)$& $0.570 (0.03)$\\
	&$40\%$	& \bestb{0.534 (0.13)}& $2.756 (2.33)$& $2.097 (1.15)$& \besta{0.251 (0.02)}& $1.650 (0.94)$& $1.239 (0.36)$&$0.615 (0.05)$  & $0.557 
(0.03)$& $0.567 (0.02)$& $0.563 (0.02)$& $0.597 (0.02)$\\
	&$50\%$	& \bestb{0.557 (0.10)}& $2.400 (1.44)$& $1.520 (0.38)$& \besta{0.268 (0.01)}& $1.612 (0.69)$& $1.144 (0.36)$&$0.665 (0.10)$  & $0.580 
(0.03)$& $0.585 (0.03)$& $0.583 (0.03)$& $0.636 (0.02)$\\ \midrule
\multirow{5}{*}{\rotatebox[origin=c]{90}{SIM-Overlapped}}&	$10\%$	& \besta{0.305 (0.02)}& $0.693 (0.31)$& $1.510 (0.97)$& \bestb{0.307 (0.02)}& 
$1.373 (0.63)$& $2.168 (0.20)$&$0.554 (0.14)$ & $0.395 (0.03)$& $0.428 (0.06)$& $0.385 (0.01)$& $0.427 (0.01)$\\
	&$20\%$	& \bestb{0.368 (0.03)}& $1.562 (0.45)$& $1.881 (0.50)$& \besta{0.293 (0.01)}& $2.702 (1.28)$& $1.837 (0.37)$&$0.608 (0.08)$ & $0.467 
(0.02)$& $0.532 (0.07)$& $0.440 (0.02)$& $0.502 (0.01)$\\
	&$30\%$	& \bestb{0.472 (0.04)}& $1.825 (0.55)$& $2.209 (0.64)$& \besta{0.294 (0.03)}& $5.101 (1.99)$& $1.568 (0.61)$&$0.586 (0.15)$ & $0.532 
(0.02)$& $0.521 (0.03)$& $0.508 (0.01)$& $0.557 (0.01)$\\
	&$40\%$	& $0.549 (0.04)$& $2.372 (0.54)$& $2.597 (0.73)$& \besta{0.322 (0.01)}& $4.569 (1.72)$& $1.320 (0.40)$&$0.687 (0.11)$ & $0.546 
(0.02)$& $0.556 (0.03)$& \bestb{0.541 (0.03)}& $0.593 (0.02)$\\
	&$50\%$	& $0.641 (0.06)$& $2.269 (0.44)$& $2.247 (0.60)$& \besta{0.298 (0.02)}& $5.762 (3.34)$& $1.174 (0.25)$&$0.815 (0.12)$ & $0.563 
(0.03)$& $0.576 (0.02)$& \bestb{0.560 (0.03)}& $0.618 (0.02)$\\ \midrule
\multirow{5}{*}{\rotatebox[origin=c]{90}{SIM-Mixed}}&$10\%$	& \besta{0.282 (0.01)}& $0.443 (0.11)$& $0.448 (0.11)$& $0.290 (0.01)$& $0.951 
(0.35)$& $2.032 (0.46)$& $0.414 (0.12)$ & $0.358 (0.01)$& $0.418 (0.06)$& $0.359 (0.01)$& $0.355 (0.01)$\\
&$20\%$	& \bestb{0.351 (0.02)}& $0.857 (0.52)$& $1.325 (0.79)$& \besta{0.286 (0.01)}& $1.062 (0.38)$& $1.782 (0.44)$& $0.462 (0.08)$ & $0.413 
(0.02)$& $0.476 (0.06)$& $0.409 (0.01)$& $0.428 (0.01)$\\
&$30\%$	& \bestb{0.396 (0.02)}& $1.368 (0.74)$& $1.524 (0.64)$& \besta{0.278 (0.01)}& $1.693 (0.56)$& $1.627 (0.54)$& $0.483 (0.07)$ & $0.454 
(0.02)$& $0.464 (0.04)$& $0.449 (0.01)$& $0.468 (0.01)$\\
&$40\%$	& \bestb{0.449 (0.03)}& $1.100 (0.61)$& $1.188 (0.59)$& \besta{0.277 (0.02)}& $1.609 (0.43)$& $1.456 (0.34)$& $0.483 (0.05)$ & $0.478 
(0.02)$& $0.504 (0.04)$& $0.478 (0.01)$& $0.508 (0.02)$\\
&$50\%$	& \bestb{0.492 (0.03)}& $1.364 (0.59)$& $1.513 (0.67)$& \besta{0.265 (0.01)}& $1.972 (0.86)$& $1.366 (0.29)$& $0.562 (0.04)$ & $0.501 
(0.01)$& $0.515 (0.02)$& $0.499 (0.02)$& $0.546 (0.02)$\\ \bottomrule
\end{tabular} }
\label{tab:db_sim}
\end{table*}

\begin{figure*}[htb!]
\begin{minipage}[b]{2\columnwidth}
\centering
\subfloat[WD-EM]{\includegraphics[trim=0mm 0mm 0mm 
8mm,clip,width=0.220\columnwidth,scale=0.2,keepaspectratio]{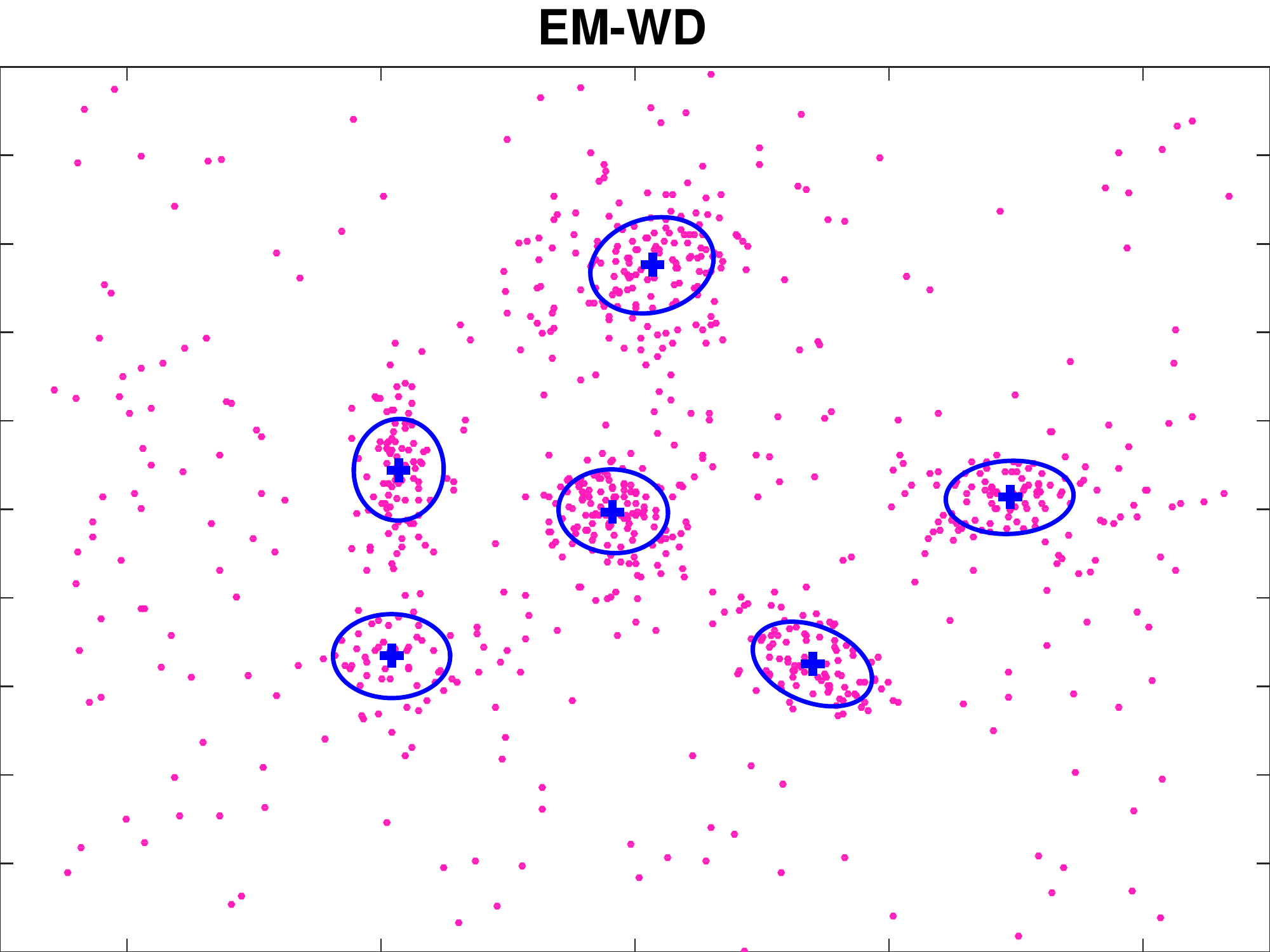}} \qquad
\subfloat[FWD-EM]{\includegraphics[trim=0mm 0mm 0mm 
8mm,clip,width=0.220\columnwidth,scale=0.2,keepaspectratio]{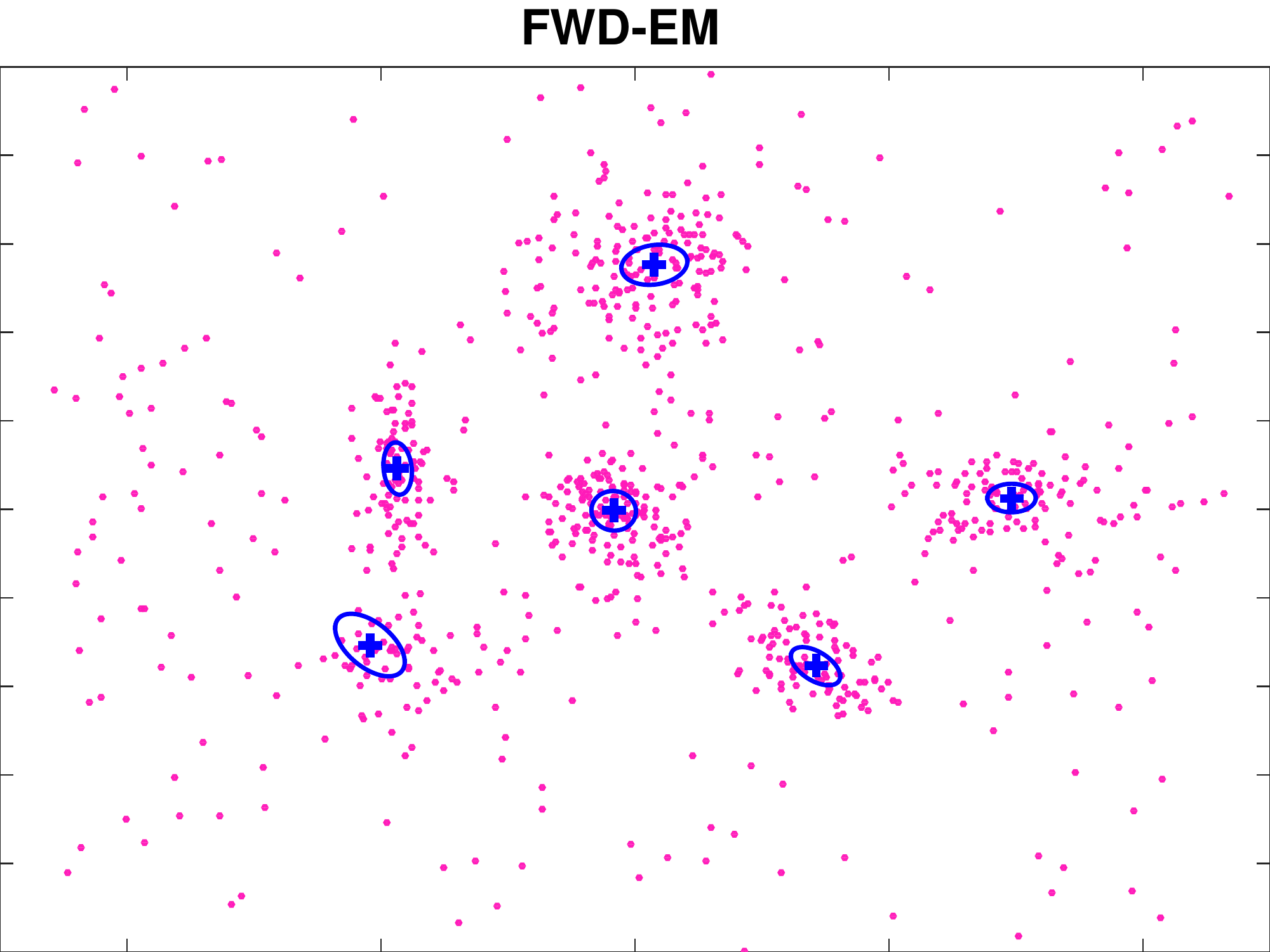}} \qquad
\subfloat[GMM]{\includegraphics[trim=0mm 0mm 0mm 
8mm,clip,width=0.220\columnwidth,scale=0.2,keepaspectratio]{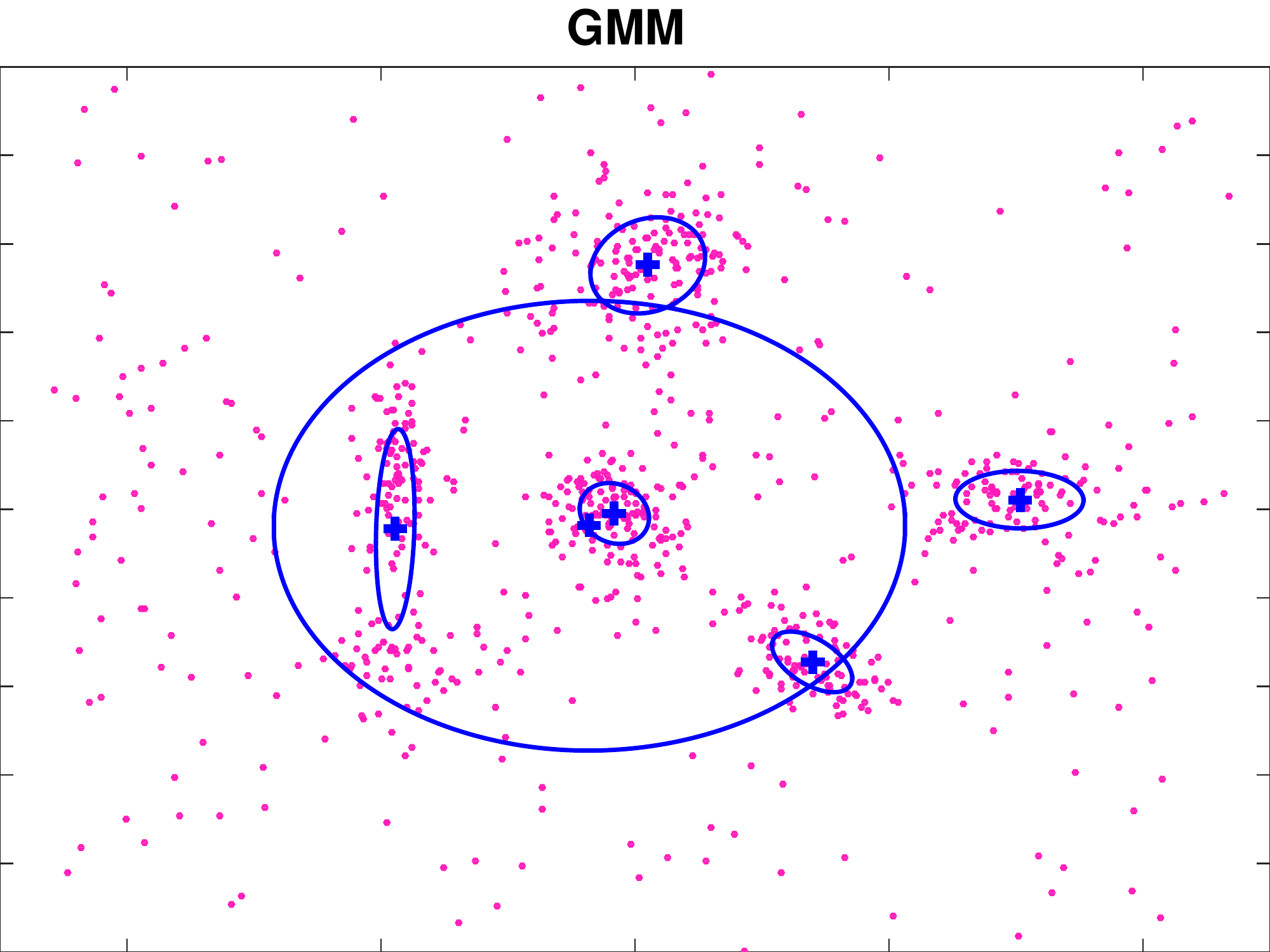}} \qquad
\subfloat[GMM+U]{\includegraphics[trim=0mm 0mm 0mm 
8mm,clip,width=0.220\columnwidth,scale=0.2,keepaspectratio]{SIM-Mixed_Algo_EM-WD_N_300.pdf}}\\ \qquad
\end{minipage}
\vspace{1.0em}
\begin{minipage}[b]{2\columnwidth}
\centering
\subfloat[FM-uMST ]{\includegraphics[trim=0mm 0mm 0mm 
8mm,clip,width=0.220\columnwidth,scale=0.2,keepaspectratio]{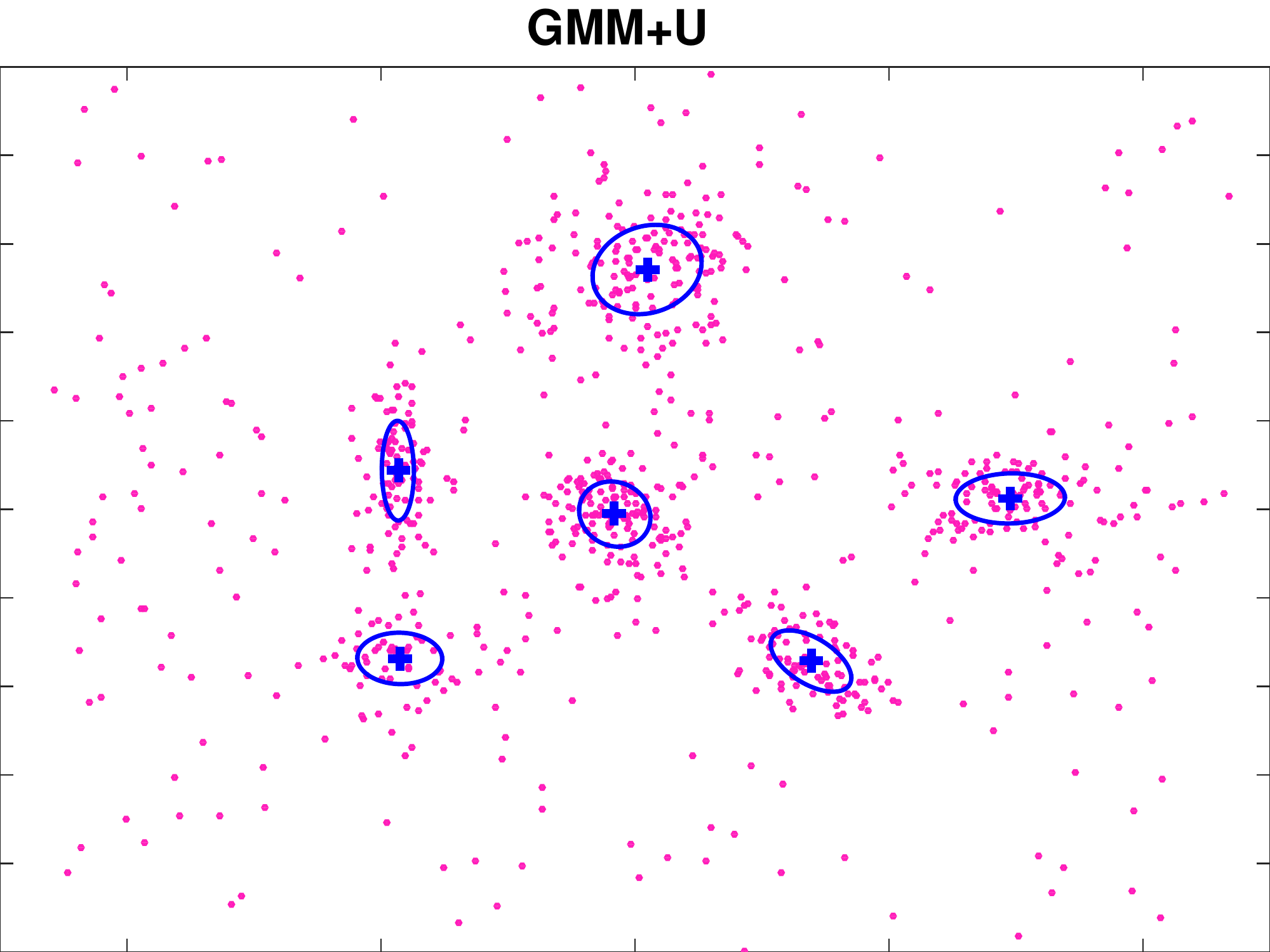}} \qquad
\subfloat[IGMM]{\includegraphics[trim=0mm 0mm 0mm 
8mm,clip,width=0.220\columnwidth,scale=0.2,keepaspectratio]{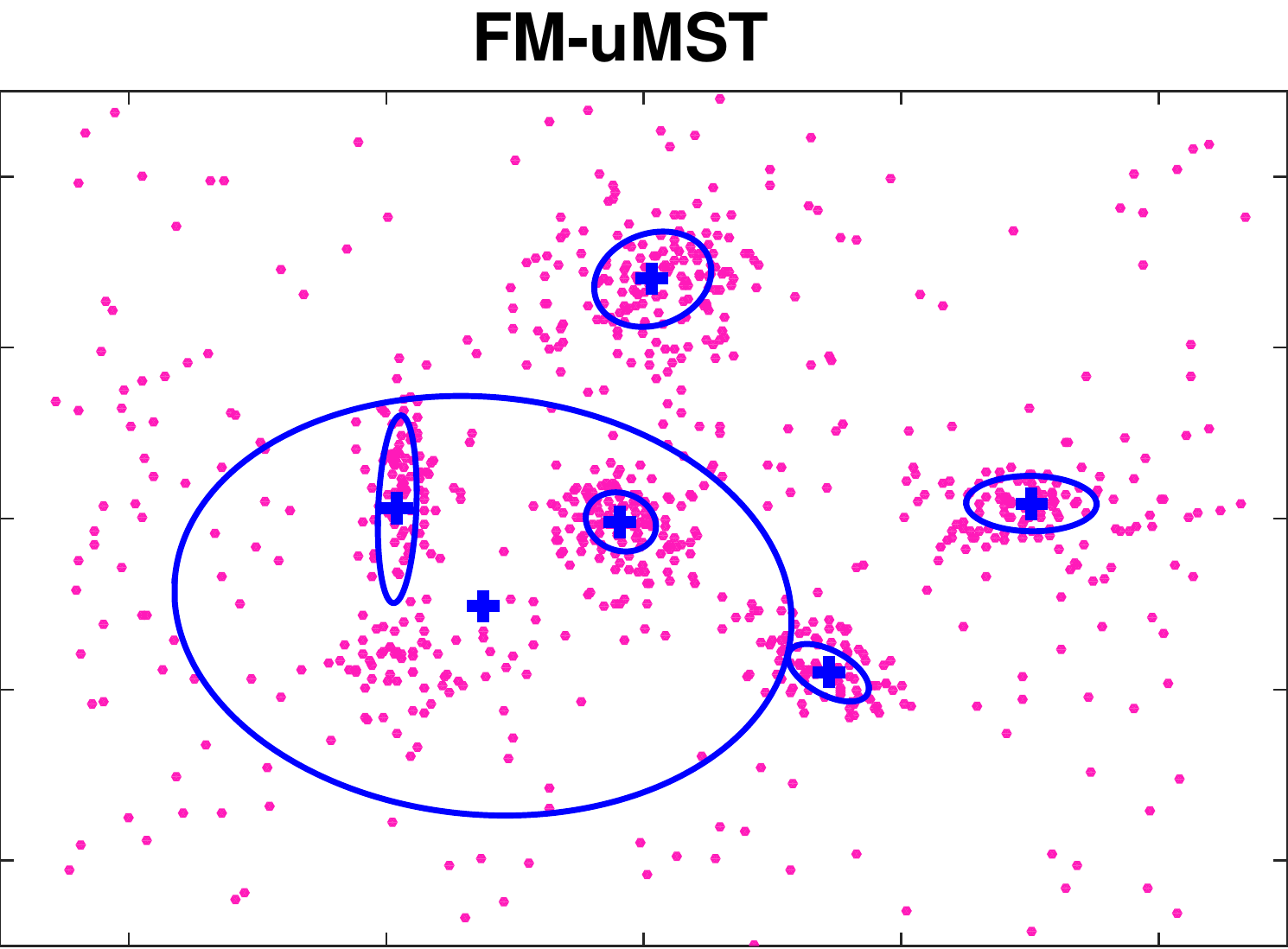}} \qquad
\subfloat[$\text{I}^2$GMM]{\includegraphics[trim=0mm 0mm 0mm 
8mm,clip,width=0.220\columnwidth,scale=0.2,keepaspectratio]{SIM-Mixed_Algo_GMM_N_300.pdf}} \qquad
\subfloat[Ground Truth]{\includegraphics[trim=0mm 0mm 0mm 
8mm,clip,width=0.220\columnwidth,scale=0.2,keepaspectratio]{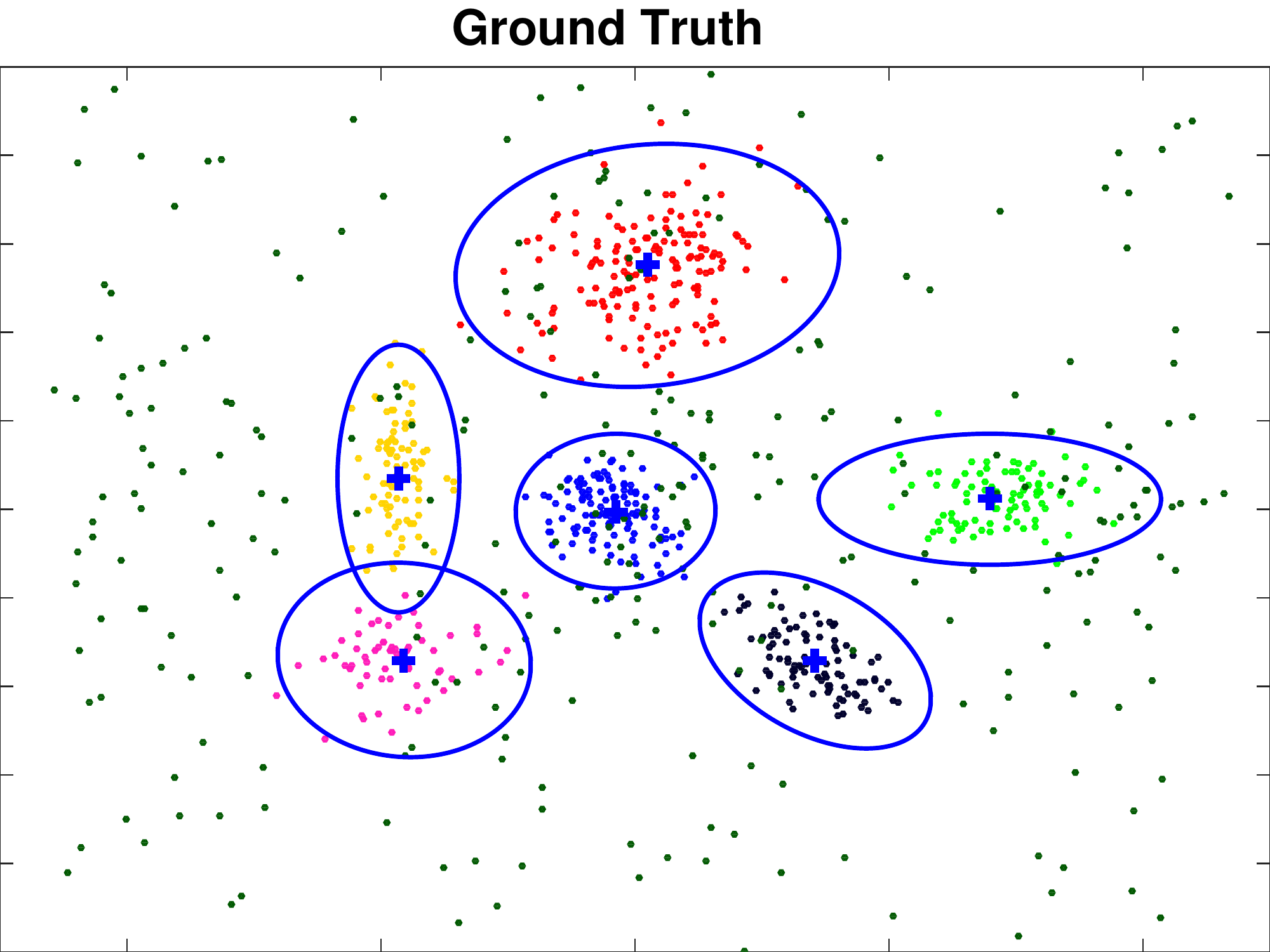}} \\ \hfill
\end{minipage}
\caption{\label{fig:sim_mixed} Results obtained by fitting mixture models to the SIM-Mixed data in the presence of 50\% outliers (see 
\tref{tab:db_sim}). }
\end{figure*} 
To quantitatively evaluate all the tested methods, we chose to use the Davies-Bouldin (DB) index \cite{davies1979cluster}:
\begin{equation}
\label{eq:DB}
 DB = \dfrac{1}{K}\sum\limits_{k=1}^{K} R_{k},
\end{equation}
where
$R_{k} = \max_{k,k\neq l} \{ (S_{k} + S_{l})/d_{kl} \}$,
$S_k = n_k\inverse \sum_{\xvect\in C_k}\norm{\xvect-\muvect_k}$ is the cluster scatter, $n_k$ is the number of samples in cluster $k$, $\muvect_k$ is the cluster center, and $d_{kl}= 
\norm{\muvect_k-\muvect_l}$. A low value of the DB index means that the clusters are far from each other with respect to their scatter, and therefore 
the discriminative power is higher. Since the algorithms are randomly initialized, we repeat each experiment 20 times and compute the mean and 
standard deviation of the DB index for each experiment. \tref{tab:db_realdataset} summarizes the results obtained with the MNIST, WAV, BCW, and Letter 
Recognition datasets. The proposed WD-EM method yields the best results for the WAV and BCW data, while the I$^2$GMM method yields the best results for the 
MNIST data. It is interesting to notice that the non-parametric methods K-means, NCUT and HAC yield the best and second best results for the Letter 
Recognition data. 

\addnote[f1-measure]{3}{For completeness we also provide the micro $\text{F}_1$ scores (also used in \cite{yerebakan2014infinite}) obtained with the MNIST, WAV, BCW and Letter 
Recognition datasets in \tref{tab:f1_realdataset}. Based on this classification score, the proposed WD-EM method yields the best results for the WAV 
and BCW data, while the I$^2$GMM yields the best results for the Letter Recognition data, and the IGMM method yields the best results for the MNIST 
data. This comparison also shows that I$^2$GMM, GMM and GMM+U yield similar scores.}

\addnote[sim-discussion]{1}{
An interesting feature of the proposed weighted-data clustering algorithms is their robustness in finding good clusters in the presence of outliers.  To illustrate this ability we ran a large number 
of experiments by adding outliers, drawn from a uniform distribution, to the four simulated datasets, \eg  
Table~\ref{tab:db_sim} and \fref{fig:sim_mixed}. A comparison between WD-EM, FWD-EM, and the state-of-art clustering techniques 
mentioned above, with different percentages of outliers, is provided. As it 
can be easily observed in these tables, GMM+U performs extremely well in the presence of outliers, which is not surprising since the simulated outliers are drawn from a uniform distribution. Overall, the proposed WD-EM method is the second best performing 
method. Notice the very good performance of the Ncut method for the SIM-overlapped data. Among all these methods, only GMM+U and WD-EM offer the possibility to characterize the outliers using two very different strategies.The GMM+U model simply pulls them in an \textit{outlier class} based on the posterior probabilities. 
The WD-EM algorithm iteratively updates the posterior probabilities of the weights, and the final posteriors, \eref{newEW}, allow to implement a simple outlier detection mechanism. Another important remark is that WD-EM systematically outperforms FWD-EM, which fully justifies the proposed weighted-data model. \fref{fig:sim_mixed} shows results of fitting  the mixture models to SIM-mixed data drawn from a Gaussian mixture and contaminated with 50\% outliers drawn from a uniform distribution. These plots show that GMM, IGMM, and I$^2$GMM find five components corresponding to data clusters while they also fit a component onto the outliers, roughly centered on the data set. 
}

\section{Audio-Visual Clustering}
\label{sec:avexperiment}

\begin{figure}[b!]
\centering
\includegraphics[width=.6\linewidth,trim=1mm 1mm 1mm 1mm,clip=,keepaspectratio]{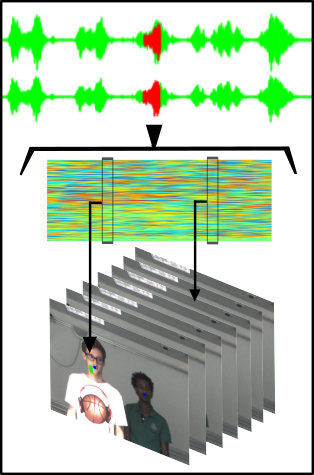}
\caption{\label{fig:avsegment} Audio-visual data acquisition and alignment. \textit{Top}: left- and right-microphone signals. A temporal segment of 0.4~s is outlined in red. \textit{Middle}: Binaural spectrogram that corresponds to the outlined segment. This spectrogram is composed of 50 binaural vectors, each one being associated with an audio frame (shown as a vertical rectangle). \textit{Bottom}: video frames associated with a segment. A sound-source direction of arrival (DOA) is extracted from each binaural vector and mapped onto the image plane, hence each green dot in the image plane corresponds to a DOA.}
\end{figure}

In this section we illustrate the effectiveness of our method to deal with audio-visual data which belong to the heterogenous type of data, \ie gathered with different sensors, having different noise statistics, and different sources of errors. The challenges of clustering audio-visual data were enumerated in \sref{sec:intro}.
Prior to clustering one needs to represent audio and visual observations in the same Euclidean space, \eg \fref{fig:avsegment}. Without loss of generality we 
adopt the sound-source localization method of \cite{deleforge:hal-01112834} that performs 2D direction of arrival (DOA) estimation followed by mapping the estimated sound-source direction onto the 
image plane: a DOA estimate therefore corresponds to a pixel location in the image plane. To find visual features, we use an upper-body detector \cite{ferrari2008progressive} that provides an approximate localization of human heads, followed 
by lip localization using facial landmark detection \cite{zhu2012face}. The rationale of combining upper-body detection with facial landmark localization is that, altogether this yields a detection 
and localization algorithm that is much more robust to head pose than the vast majority of face detection methods. 

\begin{figure*}[p]
\begin{minipage}[c]{1\columnwidth}
\centering
\begin{tabular}{cccc} 
Data \& Active speaker & WD-EM & GMM+U & FM-uMST \\
\includegraphics[trim=1mm 62mm 1mm 62mm,clip,scale=0.18]{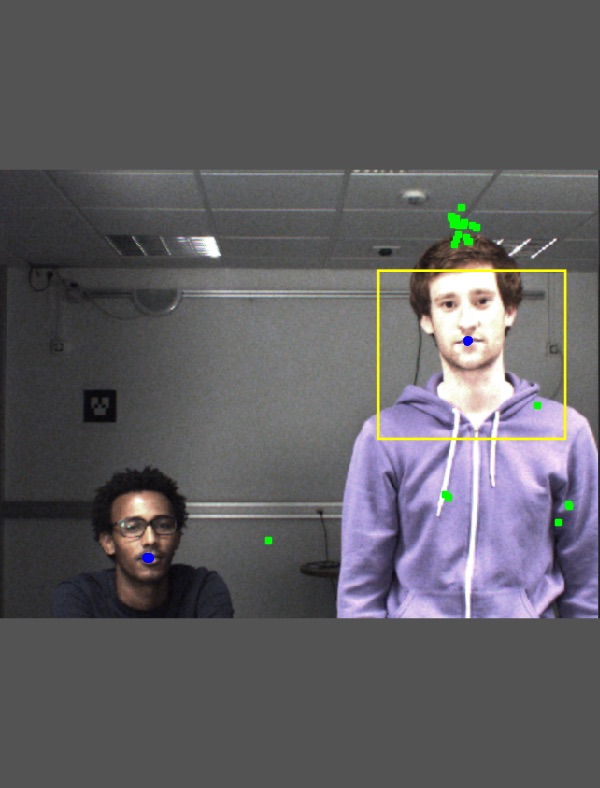} & 
\includegraphics[trim=1mm 62mm 1mm 62mm,clip,scale=0.18]{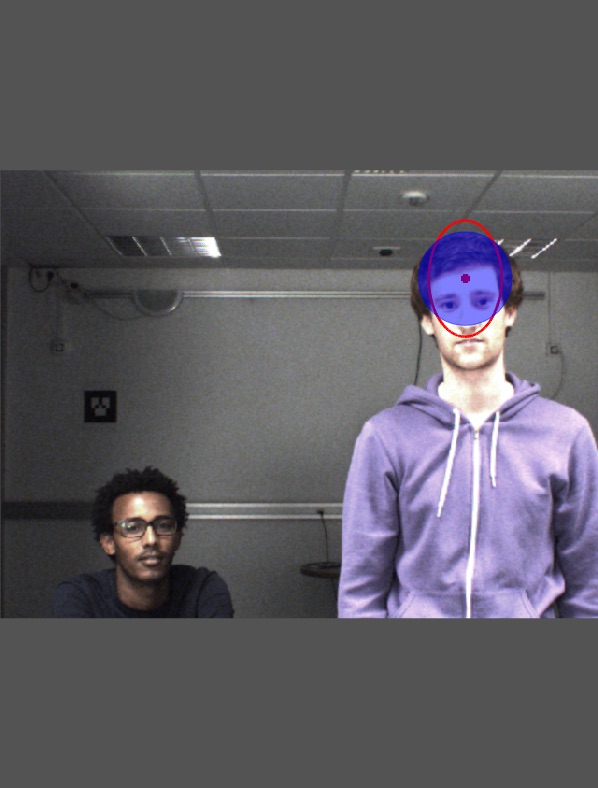} & 
\includegraphics[trim=1mm 62mm 1mm 62mm,clip,scale=0.18]{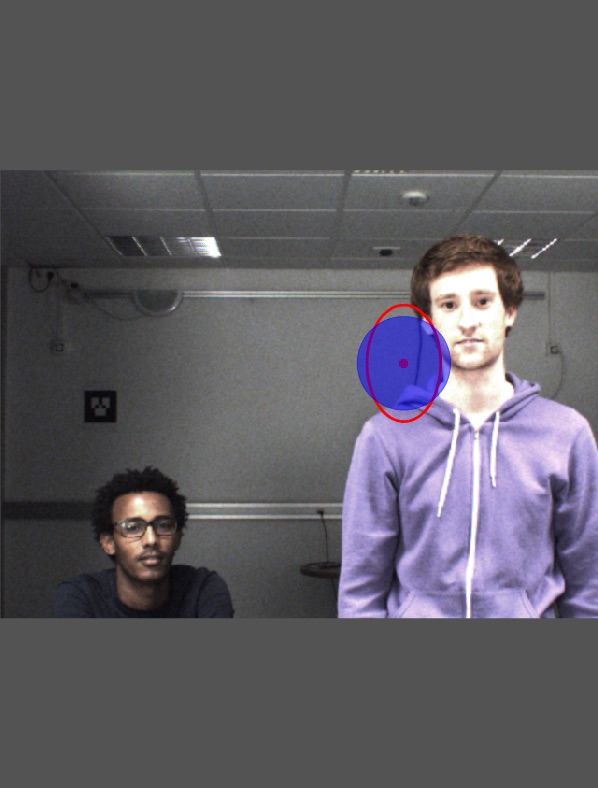} & 
\includegraphics[trim=1mm 62mm 1mm 62mm,clip,scale=0.18]{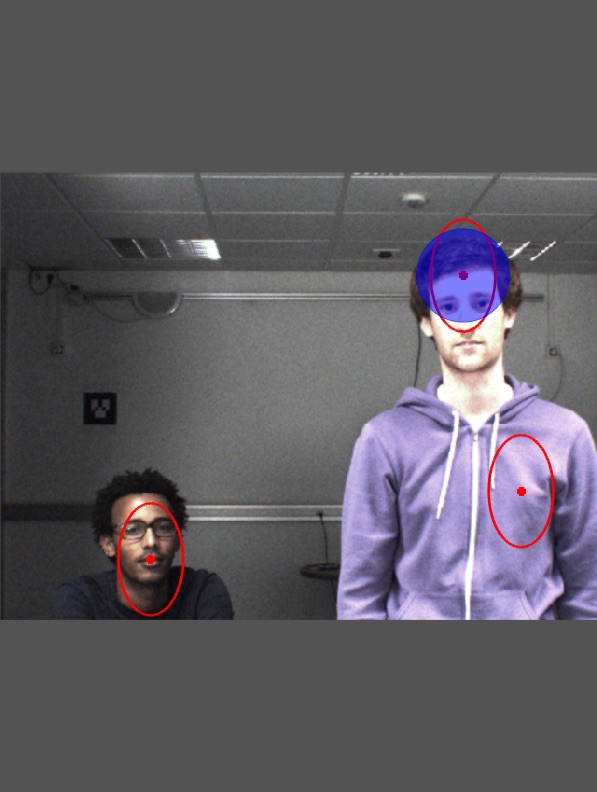}  \\
FS-A & Correct & Correct & Correct \\
\includegraphics[trim=1mm 62mm 1mm 62mm,clip,scale=0.18]{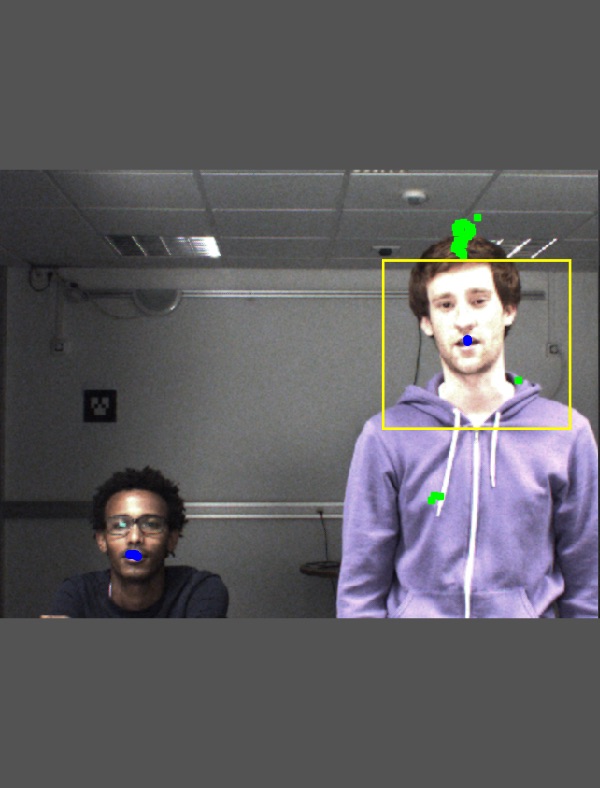} & 
\includegraphics[trim=1mm 62mm 1mm 62mm,clip,scale=0.18]{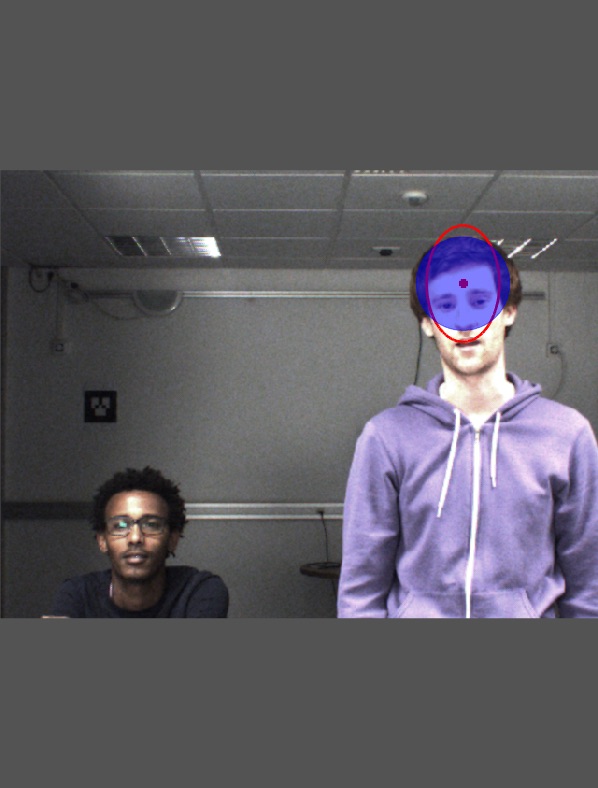} & 
\includegraphics[trim=1mm 62mm 1mm 62mm,clip,scale=0.18]{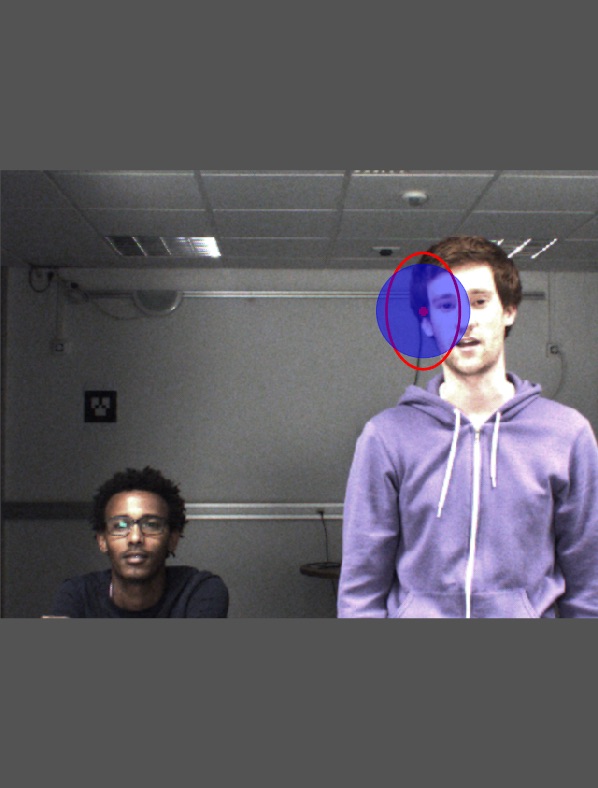} & 
\includegraphics[trim=1mm 62mm 1mm 62mm,clip,scale=0.18]{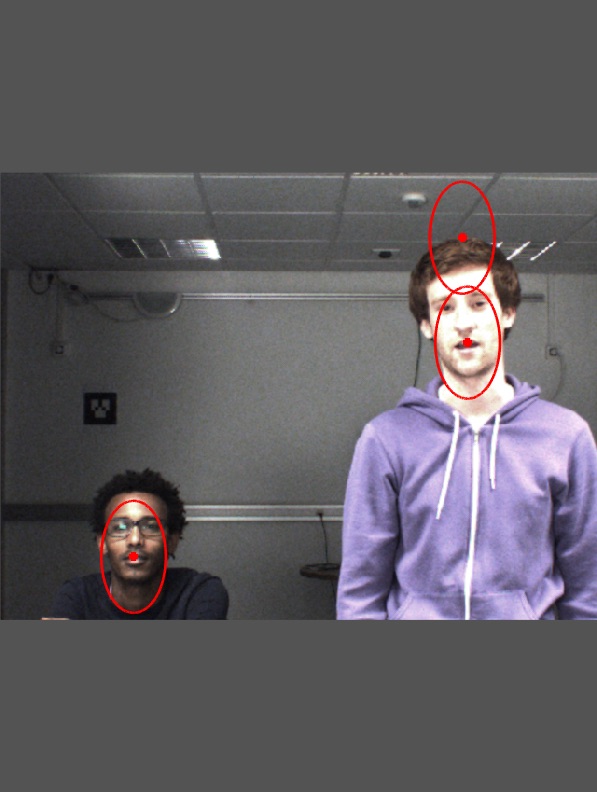}  \\
FS-B & Correct & Correct & Incorrect \\
\includegraphics[trim=1mm 1mm 0mm 0mm,clip,scale=0.3]{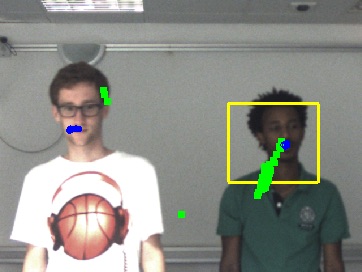} & 
\includegraphics[trim=1mm 1mm 0mm 0mm,clip,scale=0.3]{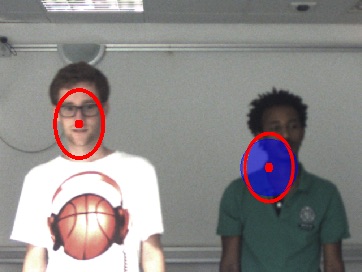} & 
\includegraphics[trim=1mm 1mm 0mm 0mm,clip,scale=0.3]{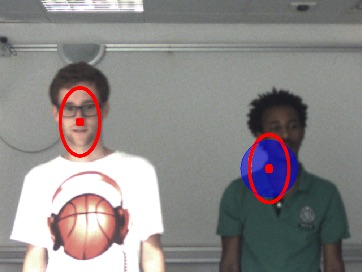} & 
\includegraphics[trim=1mm 1mm 0mm 0mm,clip,scale=0.3]{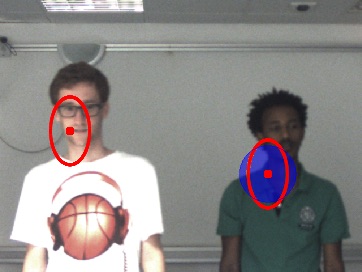}  \\
MS-A & Correct & Correct & Correct \\
\includegraphics[trim=1mm 1mm 0mm 0mm,clip,scale=0.3]{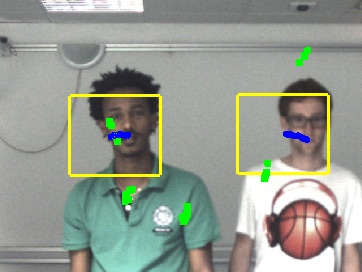} & 
\includegraphics[trim=1mm 1mm 0mm 0mm,clip,scale=0.3]{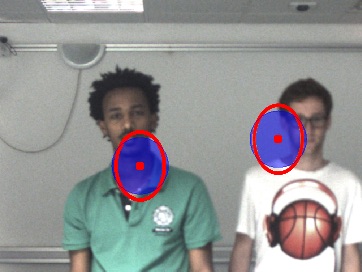} & 
\includegraphics[trim=1mm 1mm 0mm 0mm,clip,scale=0.3]{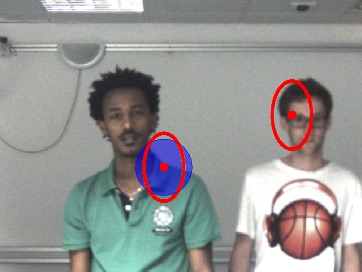} & 
\includegraphics[trim=1mm 1mm 0mm 0mm,clip,scale=0.3]{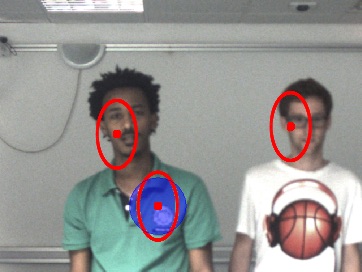}  \\
MS-B & Correct & Incorrect & Incorrect \\
\includegraphics[trim=1mm 62mm 1mm 62mm,clip,scale=0.18]{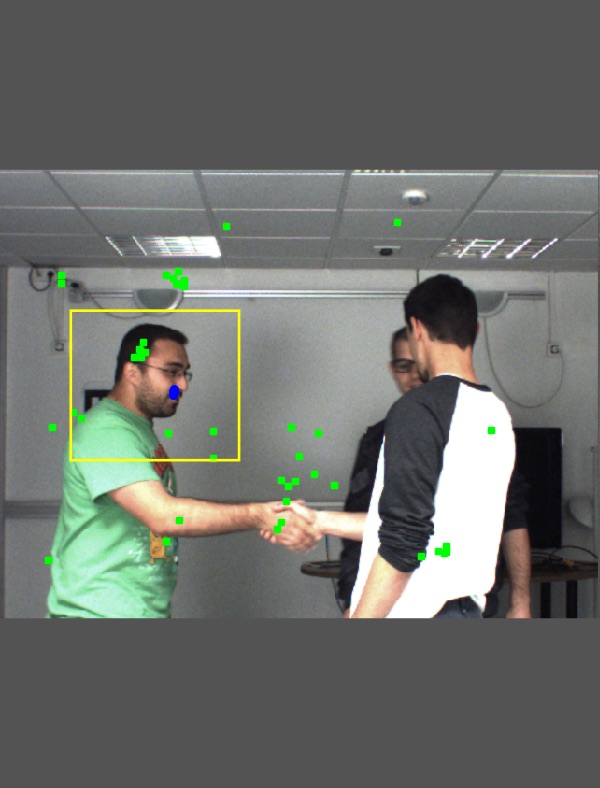} & 
\includegraphics[trim=1mm 62mm 1mm 62mm,clip,scale=0.18]{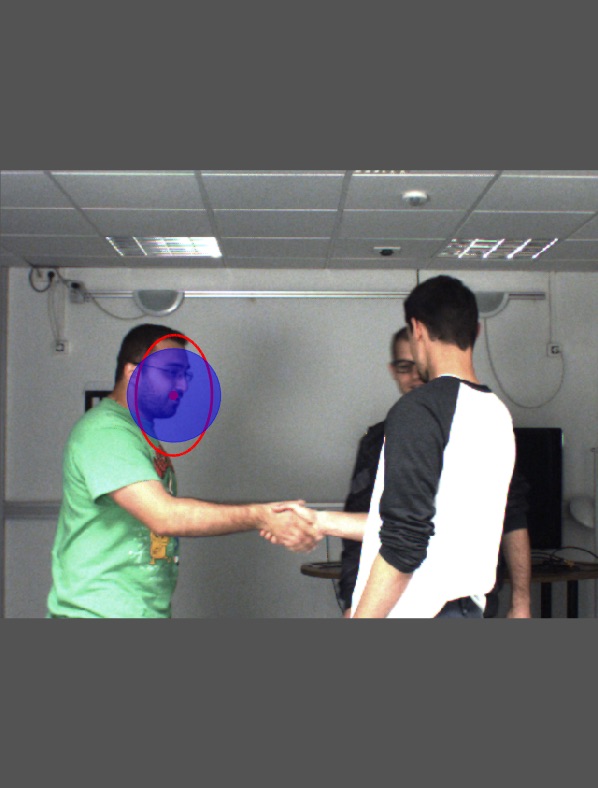} & 
\includegraphics[trim=1mm 62mm 1mm 62mm,clip,scale=0.18]{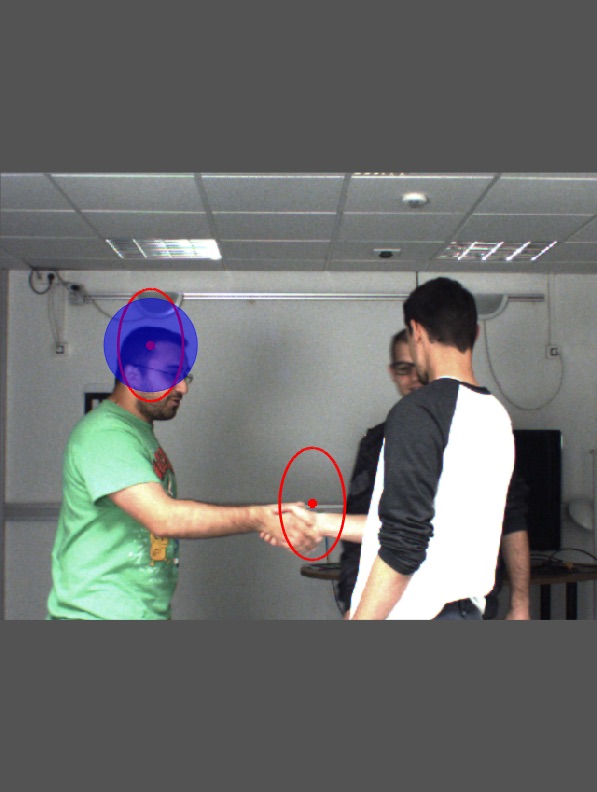} & 
\includegraphics[trim=1mm 62mm 1mm 62mm,clip,scale=0.18]{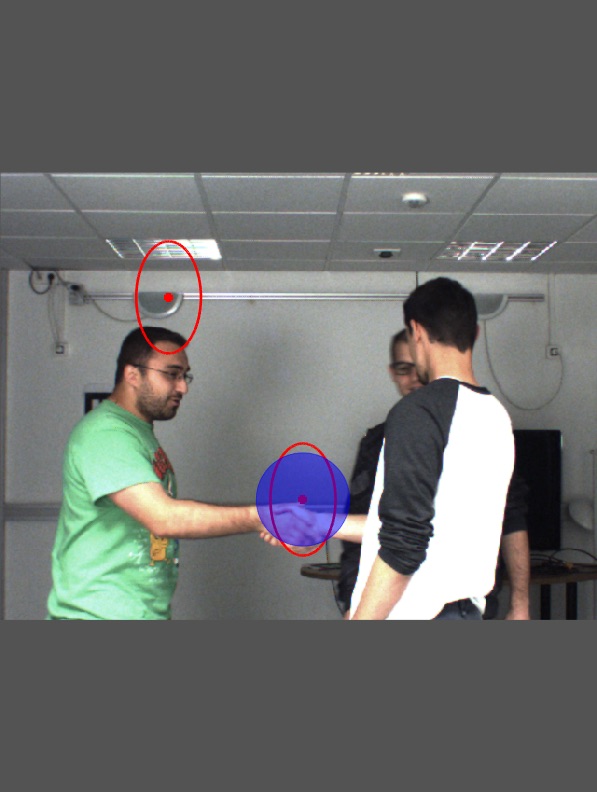}  \\
CP-A & Correct & Correct & Incorrect \\
\includegraphics[trim=1mm 62mm 1mm 62mm,clip,scale=0.18]{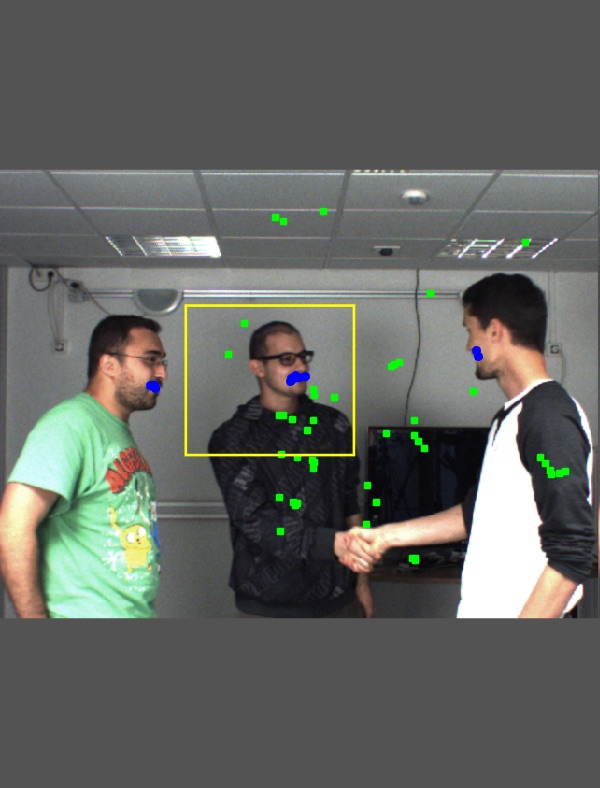} & 
\includegraphics[trim=1mm 62mm 1mm 62mm,clip,scale=0.18]{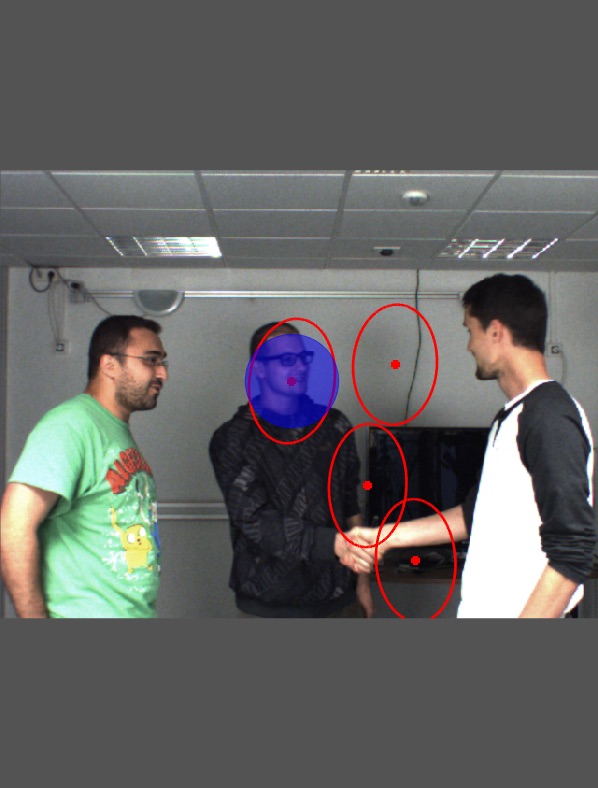} & 
\includegraphics[trim=1mm 62mm 1mm 62mm,clip,scale=0.18]{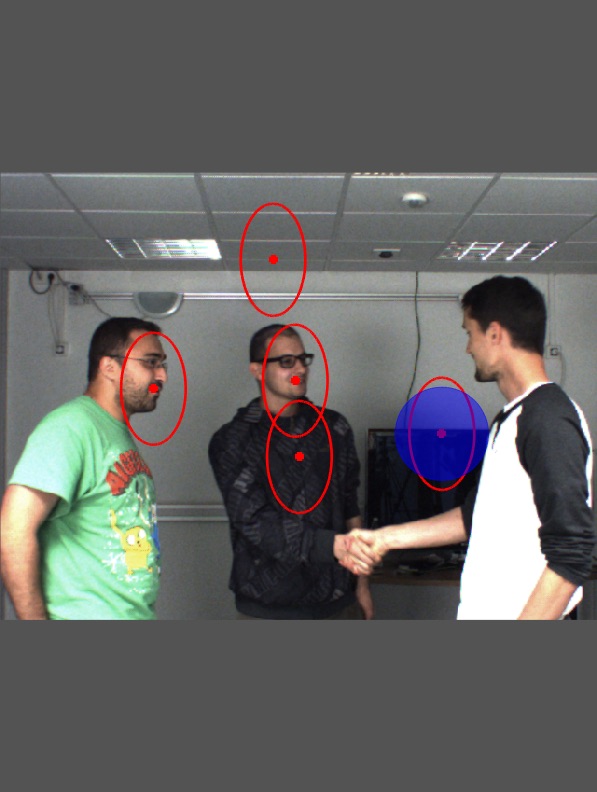} & 
\includegraphics[trim=1mm 62mm 1mm 62mm,clip,scale=0.18]{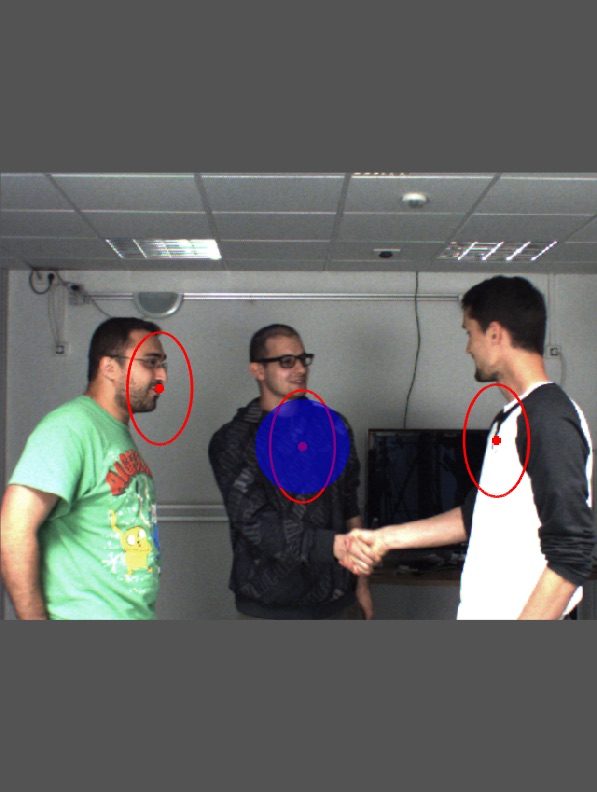}  \\
CP-B & Correct & Incorrect & Correct \\
\end{tabular}
\end{minipage}
\caption{Results obtained on the \textit{fake speaker} (FS), \textit{moving speaker} (MS) and \textit{cocktail party} (CP) sequences. The first 
column shows the audio (green) and visual (blue) observations, as well as a yellow bounding box that shows the ground-truth active speaker.  The 
second, third and fourth columns show the mixture components obtained with the WD-EM, GMM+U and FM-uMST methods, respectively. The blue disks mark components 
that correspond to correct detections of active speakers, namely whenever there is an overlap between a component and the ground-truth bounding box.}
\label{fig:fs}
\end{figure*}

Let $\Avect=\{\avect_{j}\}_{j=1}^{n_a}\in\mathbb{R}^2$ and $\Vvect=\{\vvect_{j}\}_{j=1}^{n_v}\in \mathbb{R}^2$ denote the set of auditory and visual observations respectively.
To initialize the weight variables, we use\eref{eqn:weightformula} in the following way. An auditory sample is given a high initial weight if it has many visual samples as neighbors, or $w_{a_i} = 
\sum_{\vvect_j\in\Vvect} \exp (-d^2(\avect_i,\vvect_j)/\sigma)$. Visual weights are initialized in an analogous way, $w_{v_i} = 
\sum_{\avect_j\in\Avect} \exp (-d^2(\vvect_i,\avect_j)/\sigma)$. As 
illustrated below, this \textit{cross-modal} weighting scheme favors clusters composed of both auditory and visual observations. 
We recorded three sequences:
\begin{itemize}
\item The \textit{fake speaker} (FS) sequence, \eg first and second rows of \fref{fig:fs}, consists of two persons facing the camera and the microphones. 
While the person onto the right emits speech signals 
(counting from ``one" to ``ten") the person onto the left performs fake lip, facial, and head movements as he would speak.
\addnote[moving-speakers]{1}{
\item The \textit{moving speakers} (MS) sequence, \eg third and fourth rows of \fref{fig:fs}, consists of two persons that move around while they are always 
facing the cameras and microphones. The persons take speech turns but there is a short overlap between the two auditory signals.
}
\item  The \textit{cocktail party} (CP) sequence, \eg fifth and sixth rows of \fref{fig:fs}, consists of four persons engaged in an informal dialog. The 
persons wander around and turn their heads towards the active speaker; occasionally two persons 
speak simultaneously. Moreover the speakers do not always face the camera, hence face and lip detection/localization are unreliable. 
\end{itemize}
\addnote[av-data]{1}{
The visual data are gathered with a single camera and the auditory data are gathered with two microphones plugged into the ears of an acoustic dummy 
head, referred to as \textit{binaural audition}. The visual data are recorded at 25 video frames per second (FPS). The auditory data are gathered and 
processed in the following way. First, the short-time Fourier transform (STFT) is applied to the left- and right-microphone signals which are sampled 
at 48~KHz. Second, the left and right spectrograms thus obtained are combined to yield a binaural spectrogram from which a sound-source DOA is 
estimated. A spectrogram composed of 512 frequency bins is obtained by applying the STFT over a sliding window of width $0.064$~s and shifted along 
the signal with $0.008$~s hops. An audio frame, or 512 frequency bins, is associated with each window, hence there are 125 audio frames per second 
(with 0.056~ms overlap between consecutive frames). Both the visual and audio frames are further grouped into temporal segments of width 0.4~s, hence 
there are 10 visual frames and 50 audio frames in each segment. 
}

As already mentioned, we follow the method of  \cite{deleforge:hal-01112834} to extract a sound-source DOA from each audio frame. In order to increase the robustness of audio localization, a voice activity detector (VAD) \cite{sohn1999statistical} is first applied to each frame, such that not all the frames have DOA estimates associated with them. On an average there are 40 audio DOA observations per segment. 
The FS sequence contains 28 segments, the MS sequence contains  43 segments, while the CP sequence contains 115 segments. The left hand sides of 
\fref{fig:fs} show the central frame of a segment with all the visual features (blue) and auditory features (green) available within that 
segment.

We tested the proposed WD-EM algorithm on these audio-visual data as well as the GMM+U \cite{BanfieldRaftery1993} and FM-uMST \cite{lee2014finite} algorithms. We chose to compare our method with these 
two methods for the following reasons. Firstly, all three methods are based on finite mixtures and hence they can use a model selection criterion to estimate the number of components in the mixture that best approximates clusters in the data. This is important since the number of persons and of active speakers among these persons are not known in advance. Secondly, as demonstrated in the previous section, these three methods yield robust clustering in the presence of 
outliers. 

WD-EM uses the MML criterion for model selection as described in \sref{sec:model-selection}. We implemented a model selection criterion based on BIC to optimally select the number of components with GMM+U and FM-uMST. While each algorithm yields an optimal number of components for each audio-visual segment, not all them contain a sufficient number of audio and visual observations, such that the component can be associated with an active speaker.
Therefore, we apply a simple two-step strategy, firstly to decide whether a component is \textit{audio-visual}, \textit{audio-only}, or \textit{visual-only}, and secondly to select 
the best audio-visual components. Let $n_v$ and $n_a$ be the total number of visual and audio observations in a segment. We start by assigning each observation to a component:
let $n^k_a$ and $n^k_v$ be the number of audio and visual observations associated with component $k$. Let $r_k = \min \{n^k_a, n^k_v\}/(n_a+n_v)$ measure the audio-visual relevance of a component. If $r_k \geq s$ then component $k$ corresponds to an active speaker, 
with $s$ being a fixed threshold. 

\fref{fig:fs} shows examples of applying the WD-EM, GMM+U and FM-uMST algorithms to the three sequences. One may notice that, while the visual 
observations (blue) are very accurate and form small \textit{lumps} around the moving lips of a speaker (or of a fake speaker), audio observations 
(green) are very noisy and have different statistics; this is due to the presence of reverberations (the ceiling in particular) and of other sound 
sources, such as computer fans. The ground-truth active speaker is shown with a yellow frame. The data clusters obtained by the three methods are shown 
with red ellipses. A blue disk around a cluster center designates an audio-visual cluster. Altogether, one may notice that the proposed method 
outperforms the two other methods. An interesting feature of WD-EM is that the weights give more importance to the accurate visual data (because of 
the low-variance groups of observations available with these data) and hence the audio-visual cluster centers are pulled towards the 
visual data (lip locations in these examples).
\begin{table}[t!]
\centering
\caption{The correct detection rates (CDR) obtained with the three methods for three scenarios: fake speaker (FS), moving speakers (MS), and cocktail party (CP).}
\label{tab:cdresult}
  \begin{tabular}{lcccc}
   \toprule
   Scenario &  \# Segments & WD-EM & GMM-U \cite{BanfieldRaftery1993} & FM-uMST \cite{lee2014finite}\\
   \midrule
   FS & 28 & $100.00\%$ & $100.00\%$ & $71.43\%$ \\
  \midrule
  MS & 43 &  $83.87\%$ & $61.90\%$ & $72.22\%$ \\
  \midrule
  CP & 115 &  $65.66\%$ & $52.48\%$ & $49.57\%$ \\
  \bottomrule
  \end{tabular}
\end{table}

To further quantify the performance of the three methods, we carefully annotated the data. For each segment, we identified the active speaker and we 
precisely located the speaker's lips. Let 
$\xvect_g$ be the ground-truth lip location. We assign $\xvect_g$ to a component by computing the maximum responsibility\eref{eq:responsibilities} of $\xvect_g$. When $\xvect_g$ is assigned to an
audio-visual cluster, an active speaker is said to be correctly detected if the posterior probability of $\xvect_g$ is equal or greater than $1/K$, where $K$ is the number of components.
\tref{tab:cdresult} summarizes the results obtained with the three methods. 

\section{Conclusions} \label{sec:conclusion}

We presented a weighted-data Gaussian mixture model. We  derived a maximum-likelihood formulation and we devised two EM algorithms, one that uses fixed weights (FWD-EM) and another one with weights modeled as random variables (WD-EM). While the first algorithm appears to be a straightforward generalization of standard EM for Gaussian mixtures, the second one has a more complex structure. We showed that the expectation and maximization steps of the proposed WD-EM admit closed-form solutions and hence the algorithm is extremely efficient. Moreover, WD-EM performs much better than FWD-EM which fully justifies the proposed generative probabilistic model for the weights. We extended the MML-based model selection criterion proposed in \cite{figueiredo2002unsupervised} to the weighted-data Gaussian mixture model and we proposed an algorithm that finds an optimal number of components in the data.
Interestingly, the WD-EM algorithm compares favorably with several state-of-the-art parametric and non-parametric clustering methods: it performs  particularly well in the presence of a large number of outliers, \eg up to 50\% of outliers. Hence, the proposed formulation belongs to the robust category of clustering methods.

We also applied WD-EM to the problem of clustering heterogenous/multimodal data sets, such as audio-visual data. We briefly described the audio-visual fusion problem and how it may be cast into a challenging audio-visual clustering problem, \eg how to associate human faces with speech signals and how to detect and localize active speakers in complex audio-visual scenes. We showed that the proposed algorithm yields better audio-visual clustering results than two other finite-mixture models, and this for two reasons: (i)~it is very robust to noise and to outliers and (ii)~it allows a cross-modal weighting scheme. Although not implemented in this paper, the proposed model has many other interesting features when dealing with multimodal data: it enables to balance the importance of the modalities, to emphasize one modality, or to use any prior information that might be available, for example by giving high weight priors to visual data corresponding to face/lip localization. 

\bibliographystyle{IEEEtran}


\begin{thebibliography}{10}
\providecommand{\url}[1]{#1}
\csname url@samestyle\endcsname
\providecommand{\newblock}{\relax}
\providecommand{\bibinfo}[2]{#2}
\providecommand{\BIBentrySTDinterwordspacing}{\spaceskip=0pt\relax}
\providecommand{\BIBentryALTinterwordstretchfactor}{4}
\providecommand{\BIBentryALTinterwordspacing}{\spaceskip=\fontdimen2\font plus
\BIBentryALTinterwordstretchfactor\fontdimen3\font minus
  \fontdimen4\font\relax}
\providecommand{\BIBforeignlanguage}[2]{{%
\expandafter\ifx\csname l@#1\endcsname\relax
\typeout{** WARNING: IEEEtran.bst: No hyphenation pattern has been}%
\typeout{** loaded for the language `#1'. Using the pattern for}%
\typeout{** the default language instead.}%
\else
\language=\csname l@#1\endcsname
\fi
#2}}
\providecommand{\BIBdecl}{\relax}
\BIBdecl

\bibitem{McLachlanPeel2000}
G.~McLachlan and D.~Peel, \emph{Finite Mixture Models}.\hskip 1em plus 0.5em
  minus 0.4em\relax Wiley, 2000.

\bibitem{McLachlanPeel2000b}
------, ``Robust mixture modelling using the t distribution,'' \emph{Statistics
  and Computing}, vol.~10, no.~4, pp. 339--348, 2000.

\bibitem{Bishop2005}
C.~M. Bishop and M.~Svensen, ``Robust {B}ayesian mixture modelling,''
  \emph{Neurocomputing}, vol.~64, pp. 235--252, 2005.

\bibitem{Archambeau2007}
C.~Archambeau and M.~Verleysen, ``Robust {Bayesian} clustering,'' \emph{Neural
  Networks}, vol.~20, no.~1, pp. 129--138, 2007.

\bibitem{sun2010robust}
J.~Sun, A.~Kab{\'a}n, and J.~M. Garibaldi, ``Robust mixture clustering using
  {P}earson type {VII} distribution,'' \emph{Pattern Recognition Letters},
  vol.~31, no.~16, pp. 2447--2454, 2010.

\bibitem{Andrews2012}
J.~L. Andrews and P.~D. McNicholas, ``{Model-based clustering, classification,
  and discriminant analysis via mixtures of multivariate t-distributions},''
  \emph{Statistics and Computing}, vol.~22, no.~5, pp. 1021--1029, 2012.

\bibitem{ForbesWraith2014}
F.~Forbes and D.~Wraith, ``A new family of multivariate heavy-tailed
  distributions with variable marginal amounts of tailweight: application to
  robust clustering,'' \emph{Statistics and Computing}, vol.~24, no.~6, pp.
  971--984, November 2014.

\bibitem{lee2014finite}
S.~Lee and G.~McLachlan, ``Finite mixtures of multivariate skew
  t-distributions: some recent and new results,'' \emph{Statistics and
  Computing}, vol.~24, no.~2, pp. 181--202, 2014.

\bibitem{KotzNadarajah2004}
S.~Kotz and S.~Nadarajah, \emph{{Multivariate t Distributions and their
  Applications}}.\hskip 1em plus 0.5em minus 0.4em\relax Cambridge, 2004.

\bibitem{long2006spectral}
B.~Long, Z.~M. Zhang, X.~Wu, and P.~S. Yu, ``Spectral clustering for multi-type
  relational data,'' in \emph{Proceedings of the 23rd International Conference
  on Machine learning}.\hskip 1em plus 0.5em minus 0.4em\relax ACM, 2006, pp.
  585--592.

\bibitem{tseng2007penalized}
G.~Tseng, ``Penalized and weighted {K}-means for clustering with scattered
  objects and prior information in high-throughput biological data,''
  \emph{Bioinformatics}, vol.~23, no.~17, pp. 2247--2255, 2007.

\bibitem{ackerman2012weighted}
M.~Ackerman, S.~Ben-David, S.~Branzei, and D.~Loker, ``Weighted clustering,''
  in \emph{Proceedings of AAAI}, 2012.

\bibitem{feldman2012data}
D.~Feldman and L.~Schulman, ``Data reduction for weighted and outlier-resistant
  clustering,'' in \emph{Proceedings of the Twenty-Third Annual ACM-SIAM
  Symposium on Discrete Algorithms}.\hskip 1em plus 0.5em minus 0.4em\relax
  SIAM, 2012, pp. 1343--1354.

\bibitem{forbes2010weighted}
F.~Forbes, S.~Doyle, D.~Garcia-Lorenzo, C.~Barillot, and M.~Dojat, ``A weighted
  multi-sequence {Markov} model for brain lesion segmentation,'' in
  \emph{Proceedings of the International Conference on Artificial Intelligence
  and Statistics}, vol.~9, Sardinia, Italy, 2010, pp. 225--232.

\bibitem{figueiredo2002unsupervised}
M.~A.~T. Figueiredo and A.~K. Jain, ``Unsupervised learning of finite mixture
  models,'' \emph{IEEE Transactions on Pattern Analysis and Machine
  Intelligence}, vol.~24, no.~3, pp. 381--396, 2002.

\bibitem{hennig2010methods}
C.~Hennig, ``Methods for merging gaussian mixture components,'' \emph{Advances
  in Data Analysis and Classification}, vol.~4, no.~1, pp. 3--34, 2010.

\bibitem{baudry2010combining}
J.~P. Baudry, E.~A. Raftery, G.~Celeux, K.~Lo, and R.~Gottardo, ``Combining
  mixture components for clustering,'' \emph{Journal of Computational and
  Graphical Statistics}, vol.~19, no.~2, 2010.

\bibitem{melnykov2014merging}
V.~Melnykov, ``Merging mixture components for clustering through pairwise
  overlap,'' \emph{Journal of Computational and Graphical Statistics}, 2014.

\bibitem{rasmussen1999infinite}
C.~E. Rasmussen, ``The infinite {G}aussian mixture model.'' in \emph{NIPS},
  vol.~12, 1999, pp. 554--560.

\bibitem{Gorur2010}
D.~Gorur and C.~Rasmussen, ``Dirichlet process gaussian mixture models: Choice
  of the base distribution,'' \emph{Journal of Computer Science and
  Technology}, vol.~25, no.~4, pp. 653--€"664, 2010.

\bibitem{yerebakan2014infinite}
H.~Z. Yerebakan, B.~Rajwa, and M.~Dundar, ``The infinite mixture of infinite
  {G}aussian mixtures,'' in \emph{Advances in Neural Information Processing
  Systems}, 2014, pp. 28--36.

\bibitem{Wei2012}
X.~Wei and C.~Li, ``The infinite student€™ t-mixture for robust modeling,''
  \emph{Signal Processing}, vol.~92, no.~1, pp. 224€"--234, 2012.

\bibitem{schwarz1978estimating}
G.~Schwarz \emph{et~al.}, ``Estimating the dimension of a model,'' \emph{The
  Annals of Statistics}, vol.~6, no.~2, pp. 461--464, 1978.

\bibitem{celeux2001component}
G.~Celeux, S.~Chr{\'e}tien, F.~Forbes, and A.~Mkhadri, ``A component-wise {EM}
  algorithm for mixtures,'' \emph{Journal of Computational and Graphical
  Statistics}, vol.~10, no.~4, 2001.

\bibitem{lecun1998gradient}
Y.~LeCun, L.~Bottou, Y.~Bengio, and P.~Haffner, ``Gradient-based learning
  applied to document recognition,'' \emph{Proceedings of the IEEE}, vol.~86,
  no.~11, pp. 2278--2324, 1998.

\bibitem{breiman1984classification}
L.~Breiman, J.~Friedman, C.~J. Stone, and R.~A. Olshen, \emph{Classification
  and Regression Trees}.\hskip 1em plus 0.5em minus 0.4em\relax CRC Press,
  1984.

\bibitem{street1993nuclear}
W.~Street, W.~Wolberg, and O.~Mangasarian, ``Nuclear feature extraction for
  breast tumor diagnosis,'' in \emph{IS\&T/SPIE's Symposium on Electronic
  Imaging: Science and Technology}.\hskip 1em plus 0.5em minus 0.4em\relax
  International Society for Optics and Photonics, 1993, pp. 861--870.

\bibitem{frey1991letter}
P.~W. Frey and D.~J. Slate, ``Letter recognition using {H}olland-style adaptive
  classifiers,'' \emph{Machine Learning}, vol.~6, no.~2, pp. 161--182, 1991.

\bibitem{bishop2006pattern}
C.~M. Bishop, \emph{Pattern Recognition and Machine Learning}.\hskip 1em plus
  0.5em minus 0.4em\relax Springer, 2006.

\bibitem{BanfieldRaftery1993}
J.~Banfield and A.~E. Raftery, ``{Model-based Gaussian and non-Gaussian
  clustering},'' \emph{Biometrics}, vol.~49, no.~3, pp. 803--821, 1993.

\bibitem{dhillon2004kernel}
I.~S. Dhillon, Y.~Guan, and B.~Kulis, ``{Kernel K-means: spectral clustering
  and normalized cuts},'' in \emph{Proceedings of the Tenth International
  Conference on Knowledge Discovery and Data Mining}.\hskip 1em plus 0.5em
  minus 0.4em\relax ACM, 2004, pp. 551--556.

\bibitem{shi2000normalized}
J.~Shi and J.~Malik, ``Normalized cuts and image segmentation,'' \emph{IEEE
  Transactions on Pattern Analysis and Machine Intelligence}, vol.~22, no.~8,
  pp. 888--905, 2000.

\bibitem{zhao2002evaluation}
Y.~Zhao and G.~Karypis, ``Evaluation of hierarchical clustering algorithms for
  document datasets,'' in \emph{Proceedings of the Eleventh International
  Conference on Information and Knowledge Management}.\hskip 1em plus 0.5em
  minus 0.4em\relax ACM, 2002, pp. 515--524.

\bibitem{davies1979cluster}
D.~Davies and D.~Bouldin, ``A cluster separation measure,'' \emph{IEEE
  Transactions on Pattern Analysis and Machine Intelligence}, no.~2, pp.
  224--227, 1979.

\bibitem{deleforge:hal-01112834}
A.~Deleforge, R.~Horaud, Y.~Y. Schechner, and L.~Girin, ``Co-localization of
  audio sources in images using binaural features and locally-linear
  regression,'' \emph{IEEE Transactions on Audio, Speech and Language
  Processing}, vol.~23, no.~4, pp. 718--731, Apr. 2015.

\bibitem{ferrari2008progressive}
V.~Ferrari, M.-J. Marin-Jimenez, and A.~Zisserman, ``Progressive search space
  reduction for human pose estimation,'' in \emph{Proceedings of the IEEE
  Conference on Computer Vision and Pattern Recognition}, 2008.

\bibitem{zhu2012face}
X.~Zhu and D.~Ramanan, ``Face detection, pose estimation, and landmark
  localization in the wild,'' in \emph{Proceedings of IEEE Conference on
  Computer Vision and Pattern Recognition}, 2012.

\bibitem{sohn1999statistical}
J.~Sohn, N.~S. Kim, and W.~Sung, ``A statistical model-based voice activity
  detection,'' \emph{IEEE Signal Processing Letters}, vol.~6, no.~1, pp. 1--3,
  1999.

\end{thebibliography}
\hyphenation{BarcelonaTech}

\begin{IEEEbiography}[{\includegraphics[width=1in,height=1.25in,clip,keepaspectratio]{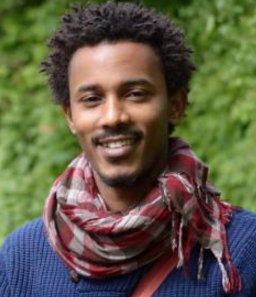}}]{Israel Dejene Gebru} received the B.Sc. degree in Computer Engineering from Addis Ababa University, Ethiopia in 2008 and the M.Sc. degree in Telecommunication Engineering from University of Trento, Italy in 2011. From 2008 to 2011, he worked for ZTE Cooperation as an Engineer. He is currently a PhD candidate in Universit\'e Joseph Fourier and works in Perception Team, at INRIA Grenoble Rh\^one-Alpes, France. His research interests include development of statistical methods and machine learning algorithms for multimodal signal analysis, speaker tracking and computer vision problems. He is particularly interested in joint processing of audio and visual data for applications in human-robot interaction.
\end{IEEEbiography}

\begin{IEEEbiography}[{\includegraphics[width=1in,height=1.25in,clip,keepaspectratio]{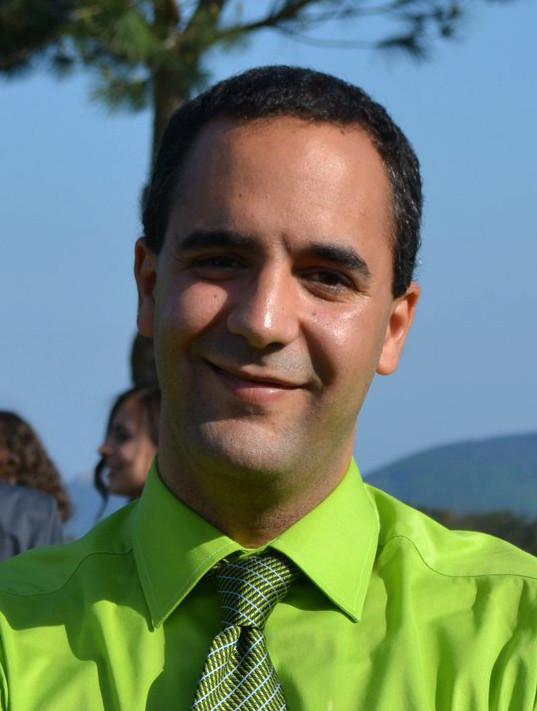}}]{Xavier Alameda-Pineda} received the M.Sc. degree in 
mathematics and telecommunications engineering from the Universitat Polit\`ecnica de Catalunya -- BarcelonaTech in 2008 and 2009 respectively, the M.Sc. degree in computer science from the 
Universit\'e Joseph Fourier and Grenoble INP in 2010, and the Ph.D. degree in mathematics/computer science from the Universit\'e Joseph Fourier in 2013. He worked towards his Ph.D. degree in the 
Perception Team, at INRIA Grenoble Rh\^one-Alpes. He currently holds a postdoctoral position at the Multimodal Human Understanding Group at University of Trento. 
His research interests are machine learning and signal processing for scene understanding, speaker diaritzation and tracking, sound source separation and behavior analysis.
\end{IEEEbiography}

\begin{IEEEbiography}[{\includegraphics[width=1in,height=1.25in,clip,keepaspectratio]{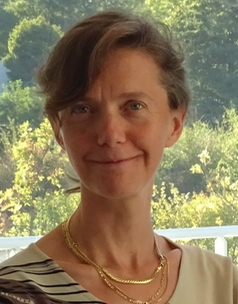}}]{Florence Forbes} 
received the B.Sc. and M.Sc.
degrees in computer science and applied
mathematics from the Ecole Nationale Sup\'erieure
d'Informatique et Math\'ematiques Appliqu\'ees de Grenoble (ENSIMAG), France, and the PhD
degree in applied probabilities from the University
Joseph Fourier, Grenoble, France. Since
1998, she has been a research scientist with INRIA, where she currently holds a position of director of research. She founded the MISTIS team and has been the team head
since 2003. Her research activities include Bayesian analysis,
Markov and graphical models, and hidden structure models.

\end{IEEEbiography}

\begin{IEEEbiography}[{\includegraphics[width=1in,height=1.25in,clip,keepaspectratio]{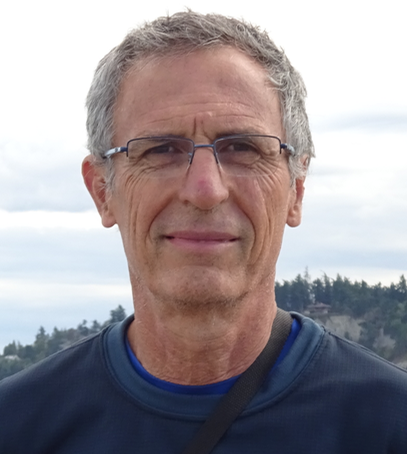}}]{Radu Horaud} 
received the B.Sc. degree in electrical engineering, the M.Sc. degree
in control engineering, and the Ph.D. degree in computer science from
the Institut National Polytechnique de Grenoble, France. 
Currently he holds a position of director of research with INRIA Grenoble, Montbonnot Saint-Martin, France, where
he is the founder and head of the PERCEPTION team. His
research interests include computer vision, machine learning, audio signal processing, 
audiovisual analysis, and robotics. He is an area editor of the
\textit{Elsevier Computer Vision and Image Understanding}, a member of
the advisory board of the \textit{Sage International Journal of Robotics
  Research}, and an associate editor of the
\textit{Kluwer International Journal of Computer Vision}. In 2013 Radu Horaud was awarded an ERC Advanced Grant for his project \textit{Vision and Hearing in Action} (VHIA).
\end{IEEEbiography}

\end{document}